\documentclass[sigconf]{acmart}

\usepackage{subcaption}
\usepackage{multirow}
\usepackage{bm}
\def\x{$\times$}
\newlength\savewidth

\AtBeginDocument{%
  \providecommand\BibTeX{{%
    \normalfont B\kern-0.5em{\scshape i\kern-0.25em b}\kern-0.8em\TeX}}}

\copyrightyear{2021}
\acmYear{2021}
\setcopyright{acmcopyright}
\acmConference[Woodstock '18]{Woodstock '18: ACM Symposium on Neural
  Gaze Detection}{June 03--05, 2018}{Woodstock, NY}
\acmPrice{15.00}
\acmDOI{10.1145/3394171.3413860}
\acmISBN{978-1-4503-7988-5/20/10}

\begin{document}

\title{Temporal Action Proposal Generation with Transformers}

\thanks{$*$ Co-first authorship: the order of first authors was randomly selected.
$\dag$ Corresponding author.}

\author{Lining Wang$^*$}
\affiliation{%
  \institution{Harbin Institute of Technology}
}

\author{Haosen Yang$^*$}
\affiliation{%
  \institution{Harbin Institute of Technology}
}

\author{Wenhao Wu$^*$}
\affiliation{%
  \institution{Department of Computer Vision Technology (VIS), Baidu Inc.}
}

\author{Hongxun Yao$^\dag$}
\affiliation{%
  \institution{Harbin Institute of Technology}
}

\author{Hujie Huang}
\affiliation{%
  \institution{Harbin Institute of Technology}
}





\begin{abstract}
Transformer networks are effective at modeling long-range contextual information and have recently demonstrated exemplary performance in the natural language processing domain.
Conventionally, the temporal action proposal generation (TAPG) task is divided into two main sub-tasks: boundary prediction and proposal confidence prediction, which rely on the frame-level dependencies and proposal-level relationships separately.
To capture the dependencies at different levels of granularity, this paper intuitively presents an unified temporal action proposal generation framework with original Transformers, called \textbf{TAPG Transformer}, which consists of a \emph{Boundary Transformer} and a \emph{Proposal Transformer}.
Specifically, the Boundary Transformer captures long-term temporal dependencies to predict precise boundary information and the Proposal Transformer learns the rich inter-proposal relationships for reliable confidence evaluation.
Extensive experiments are conducted on two popular benchmarks: ActivityNet-1.3 and THUMOS14, and the results demonstrate that TAPG Transformer outperforms state-of-the-art methods. 
Equipped with the existing action classifier, our method achieves remarkable performance on the temporal action localization task.
Codes and models will be available.


\end{abstract}




\maketitle

\section{Introduction}

With the rapid development of the mobile devices and Internet, hours of video were uploaded to the Internet every second. 
Huge video information has far exceeded the processing capacity of the conventional manual system, thus video content analysis methods have attracted increased interest from both academia and industry.

\begin{figure}[t]
\begin{center}
\includegraphics[width=0.49\textwidth]{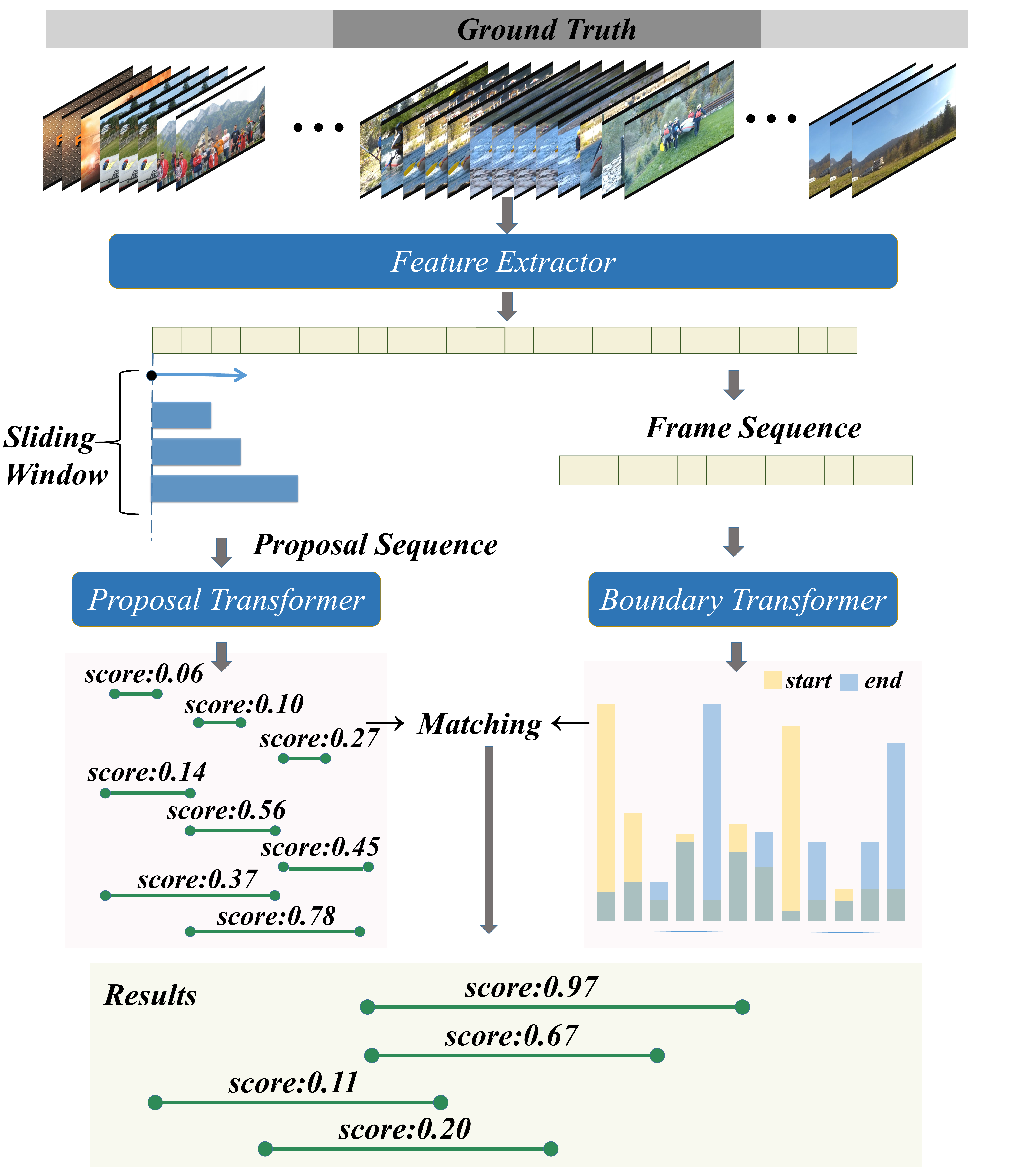}
\caption{Overview of TAPG Transformer. Given an untrimmed video, Boundary Transformer captures inter-frame relation and Proposal Transformer encodes inter-proposal relation. We further match the boundary probabilities with proposal confidence sequence to generate the final proposals.}
\label{fig:overview}
\end{center}
\end{figure}
Video temporal action detection, is one of the most active research topics in video understanding, which focuses both on classifying the action instances present in an untrimmed video and on localizing them with temporal boundaries.
Similar to object detection, temporal action detection task is composed of two sub-tasks: temporal action proposal generation (TAPG) and action recognition.
Recently, significant progress has been achieved in action recognition (\cite{two-stream,tsn,tsm,slowfast}) following the deep convolutional network paradigm. 
However, the performance of such two-stage temporal action detectors still have much room for improvement in mainstream benchmarks~\cite{thumos, activitynet}, which is largely determined by the proposal quality from TAPG.
Hence, many recent works focus on generating the high quality temporal action proposals which have flexible duration, precise boundaries and reliable confidence scores.

In general, there are two main categories in the existing TAPG methods: \emph{anchor-based} regression methods and the \emph{boundary-based} regression methods.
These two kinds of regression methods usually adopt context information in video from different aspects.
\emph{anchor-based} methods~\cite{shou2016temporal, liu2019multi, chao2018rethinking, gao2018ctap, gao2017turn, gao2020accurate, escorcia2016daps} generate action proposals based on multi-scale and dense pre-defined anchors, in this way, we can evaluate confidence scores of proposals with rich proposal-level context information. Therefore, these methods can obtain reliable confidence scores, but are inflexible and usually have imprecise boundaries.
Recently, \emph{boundary-based} methods~\cite{lin2018bsn, lin2019bmn, lin2020fast} utilize the frame-level context information around the boundaries to predict boundaries. Thus, compared with \emph{anchor-based} methods, they can generate proposals with more flexible durations and more precise boundaries. Meanwhile, they are more sensitive to noises and fail to consider the rich proposal-level context for confidence evaluation.
Based on these above analyses, we attempt to take fuller advantage of both frame-level context and proposal-level context for temporal proposal generation.

In this paper, we focus on inter-frame relation modeling and inter-proposal relation modeling for temporal proposal generation.
Inspired by the successful application of Transformers~\cite{vaswani2017attention} in sequence prediction task, we intuitively take advantage of Transformers in modeling long-range contextual information. Therefore, we propose an unified temporal action proposal generation framework with Transformers, called TAPG Transformer.
As shown in Figure~\ref{fig:overview}, our TAPG Transformer mainly contains two following modules: \emph{\textbf{Boundary Transformer}} and \emph{\textbf{Proposal Transformer}}.
\emph{Boundary Transformer} aims to locate precise action boundaries by capturing the rich long-term temporal relationships between local details and global dependencies.
To do so, the sequence of video features is provided as an input to the Transformer and output of the module is boundary probabilities.
Then, \emph{Proposal Transformer} is proposed to capture the potential inter-proposal relationships for confidence evaluation. 
Before performing inter-proposal relations, we propose a \emph{Sparse Sampling Mechanism} to generate the sparse proposal sequence instead of the densely distributed proposals which will bring imbalanced data distribution between positive/negative proposals and more computational burdens.
Then we feed them into the Proposal Transformer which discovers the complex relationships between proposals with different scales, and the Transformer outputs proposal confidences.
Finally, we take the boundary probabilities and proposal confidences as input to \emph{Fuzzy Matching} mechanism then get the matching proposals.
Followed by the post processing, we obtain the final proposal set.

In summary, our work has three main contributions as follows:
\begin{itemize}
    \item We propose a Boundary Transformer to capture the long-term frame-level dependencies for accurate temporal boundary prediction.
    \item We propose a Proposal Transformer with a sparse proposal sampling mechanism, which can learn the proposal-level context for proposal confidence evaluation. Besides, sparse sampling can significantly reduce the impact brought by the densely distributed proposals. 
    \item Extensive experiments demonstrate that our method outperforms the existing state-of-the-art methods on THUMOS14 and achieves comparable performance on ActivityNet-1.3, in both temporal action proposal generation task and temporal action detection task.
\end{itemize}

\section{Related work}
\subsection{Video Action Recognition}
Action recognition is a fundamental and important task of the video understanding area.
Currently, the end-to-end action recognition methods can be mainly divided into two types: 3D CNN-based methods and 2D CNN-based methods.
3D CNNs are intuitive spatio-temporal networks and natural extension from their 2D counterparts, which directly tackle 3D volumetric video data~\cite{c3d,i3d} but have a heavy computational cost. 
Recent studies have shown that factorizing 3D convolutional kernel into spatial (\emph{e.g.}, 1\x3\x3) and temporal components (\emph{e.g.}, 3\x1\x1) is preferable to reduce complexity as well as boost accuracy, \emph{e.g.}, P3D~\cite{p3d}, R(2+1)D~\cite{r2+1d}, S3D-G~\cite{s3d}, \emph{etc}.
Thus, 3D channel-wise convolution are also applied to video recognition in CSN~\cite{CSN} and X3D~\cite{feichtenhofer2020x3d}.
To capture temporal information with reasonable training resources, the other alternative 2D CNN based architectures are developed \emph{i.e., } TSM~\cite{tsm}, TEI~\cite{teinet}, TEA~\cite{li2020tea}, MVFNet~\cite{wu2020MVFNet}, \emph{etc}. 
These methods aim to design efficient temporal module based on existing 2D CNNs to perform efficient temporal modeling.
There is also active research on dynamic inference~\cite{wu2020dynamic}, adaptive frame sampling techniques~\cite{wu2019adaframe,wu2019multi,korbar2019scsampler}, which we think can be complementary to the end-to-end video recognition approaches.

\begin{figure*}[t]
    \centering
    \includegraphics[width=1\textwidth]{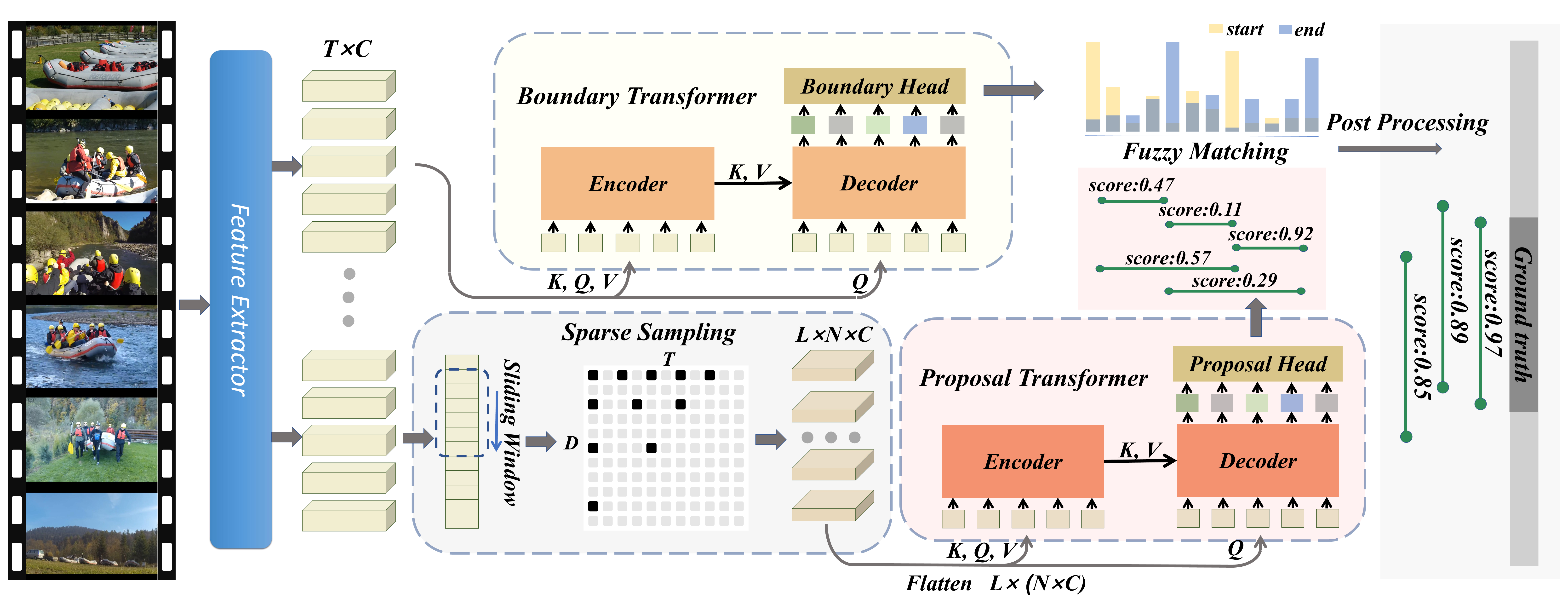}
      \caption{Illustration of the proposed TAPG Transformer framework. First we apply the feature extractor to encode video frames. The Boundary Transformer takes the feature sequence as input, and outputs boundary probability sequence. For Proposal Transformer, we apply sliding window and sparse sampling on the feature sequence to generate sparse multi-scale proposals, which are then fed into Proposal Transformer to generate confidence sequence. Finally, we construct proposals based on boundary probabilities sequence and obtain the corresponding confidence score from confidence sequence.}
    \label{fig:framework}
\end{figure*} 

\subsection{Temporal Action Proposal Generation}
Current temporal action proposal generation methods can be mainly divided into anchor-based and boundary-based methods. 
The anchor-based methods refer to the temporal boundary refinements of sliding windows or pre-defined anchors.
~\cite{oneata2014lear,thumos} use temporal sliding window and ~\cite{shou2016temporal, zhao2017temporal} use snippet-wise probability to generate candidate proposals.
TURN ~\cite{gao2017turn},  CTAP ~\cite{gao2018ctap} and Rap~\cite{gao2020accurate} adopt pre-defined anchors to generate proposals.
These methods can evaluate confidence scores of proposals with rich proposal-level context information and obtain reliable confidence scores.
Boundary-based methods generate the boundary probability sequence, then apply the Boundary Matching mechanism to generate candidate proposals.
BSN~\cite{lin2018bsn}, BMN~\cite{lin2019bmn} and BSN++~\cite{su2020bsn++} directly predict the probability of boundaries and cover flexible temporal duration of action instances.
Also, the ideas of GCN have been adopted in some works, G-TAD~\cite{xu2020g} extract temporal context and semantic context by GCN, but applies conventional temporal convolutional layers for actual implementation in temporal context extraction.
They can generate proposals with more flexible durations and more precise boundaries.
MGG~\cite{liu2019multi} and TCANet~\cite{qing2021temporal} combines a boundary-based  and anchor-based  advantages to generate proposal.
However, they have limited capability of modeling dependencies from the local context, leading to performance penalty in the complex cases.
In our work, we aim to make fuller use of inter-frame relation and inter-proposal relation for boundary prediction and the proposal-level confidence evaluation, respectively.

\subsection{Transformers in computer vision}
Transformer architectures are based on a self-attention mechanism that learns the relationships between elements of a sequence.
The breakthroughs from Transformer networks~\cite{vaswani2017attention} in language domain has sparked great interest in the computer vision community to adapt these models for vision.
ViT~\cite{dosovitskiy2020image} proposes to use NLP Transformer encoder directly on image patches.
In order to apply Transformer model, DETR~\cite{carion2020end} treats object detection as a set prediction problem. 
Transformers are also adopted for Super resolution in~\cite{yang2020learning}, Image Colorization in~\cite{kumar2021colorization}, Tracking in~\cite{xu2021transcenter, zhu2021looking, chen2021transformer}, Pose estimation in~\cite{mao2021tfpose}, \emph{etc}.
Besides, for video understanding, there are also recent approaches seek to resolve this challenge using the Transformer networks.
Video Transformer Network (VTN) that first obtain frame-wise features using 2D CNN and apply a Transformer encoder (Longformer~\cite{neimark2021video}) on top to learn temporal relationships.
TimeSformer~\cite{bertasius2021space} presents a convolution-free approach to video classification built exclusively on Transformers over space and time.
Our work is most closely related to the works which apply Transformers for temporal action proposal generation.
RTD-Net ~\cite{tan2021relaxed} adapts the Transformer-like architecture for action proposal generation in videos.
However, there are substantial differences between RTD-Net and our method.
RTD-Net uses decoder-only Transformer to exploit the global temporal context and devises a three-branch detection head for training and inference.
On the contrary, in our paper we propose to adopt two original Transformers (Boundary Transformer, Proposal Transformer) to capture both inter-frame and inter-proposal relationships for precise boundary prediction and proposal confidence evaluation, respectively.
On both two challenge benchmarks (\emph{i.e.,} THUMOS14 and ActivityNet13),
our method outperforms RTD-Net with a clear margin.


\section{Approach}

In this section, first, we present the problem definition for the temporal action proposal generation task, and then we describe the basic Transformer architecture~\cite{vaswani2017attention} that applied attention mechanisms to make global dependencies in a sequence data.
In Sec.~\ref{sec:TAPG-TR}, we present an overview of our \textbf{TAPG Transformer}. The architecture of TAPG Transformer is composed of two main parts: a \textbf{Boundary Transformer} (Sec.~\ref{sec:B-TR}) and a \textbf{Proposal Transformer} (Sec.~\ref{sec:P-TR}).
Finally, the training and inference details of our TAPG Transformer are introduced in Sec~\ref{sec:train} and Sec~\ref{sec:infer}, respectively.

\subsection{Problem Definition}
\label{sec:definition}
An untrimmed video $U$ can be denoted as a frame sequence $U=\{u_t\}_{t=1}^{l_v}$ with $l_v$ frames, where $u_t$ indicates the $t$-th RGB frame of video $U$. 
The temporal annotation set of $U$ is composed by a set of temporal action instances as $\Psi_g=\{\varphi_n^g\}_{n=1}^{N_g}$ and $\varphi_n^g = (t_{s_n}, t_{e_n})$, where $N_g$ is the number of ground-truth action instances, $t_{s_n}$ and $t_{e_n}$ are the starting and ending time of the action instance $\phi_n^g$ respectively. 
During training phase, the $\Psi_g$ is provided.
While in the testing phase, the predicted proposal set $\Psi_p$ should cover the $\Psi_g$ with high recall and high temporal overlapping.

\subsection{Preliminary: General Transformer}
\label{sec:transformer}
The original Transformer~\cite{vaswani2017attention} has an encoder-decoder structure.
The \textbf{encoder} is composed of six identical blocks, with each block having two sub-layers: a multi-head self-attention layer and a simple position-wise fully connected feed-forward layer.
Similar to the encoder, the \textbf{decoder} in the Transformer model consists of six identical blocks.
In addition to the two sub-layers in each encoder block, the decoder inserts a third sub-layer, which performs multi-head attention on the output of the corresponding encoder block. 

A key feature of the Transformer architecture is the so-called self-attention mechanism, which explicitly models the interactions between all entities of a sequence for structured prediction tasks.
Formally, lets denote a sequence of $n$ entities by $F \in \mathbb{R}^{n\times d}$, where $d$ is the embedding dimension to represent each entity. The input sequence $F$ is projected onto three learnable linear transformations to get queries $Q$, keys $K$, and values $V$.
Then, we compute the output of the self-attention mechanism as:
\begin{equation}
    Attention (Q,K,V) = softmax(\frac{QK^T}{\sqrt{d}}) \cdot V.
\end{equation}

\subsection{TAPG Transformer}
\label{sec:TAPG-TR}
Temporal action proposal generation task is generally divided into two sub-task: boundary prediction and proposal confidence evaluation.
The long-range frame-level dependencies are desirable in precise boundary prediction.
Moreover, modeling the rich inter-proposals relationships play a crucial role in confidence regression.
Motivated by the success of Transformer models in the language domain, 
similar to how a Transformer operates on a sentence, it could naturally be applied to a sequence of frames or proposals. 
Therefore, we propose to solve each task sequentially by a different Transformer network which enables capturing the long-term information and dependencies between sequence elements.
Figure~\ref{fig:framework} illustrates the architecture of our proposed framework, which called TAPG Transformer. 

Given an untrimmed video $U$ which contains $l_v$ frames, we process the input video in a regular interval $\sigma$ for reducing the computational cost. 
Then we utilize the feature extractor to encode the visual feature from video frames.
We concatenate the output of the last fully connected layer in the two-stream network~\cite{two-stream} to form the feature sequence $F = \{f_i\}^{T}_{i=1} $, where $T = l_v / \sigma$. Finally, the feature sequence $F$ is used as the input of our framework.

Our model mainly contains two main modules: \textbf{Boundary Transformer} and \textbf{Proposal Transformer}.
Boundary Transformer aims to capture the long-range dependencies between the frame-level features then output the boundary probabilities sequence.
Similar to the Boundary Transformer which learns the relationships between frames, we propose a Proposal Transformer to model the relation between proposals with different scales and evaluate the confidence of proposals.
Before performing inter-proposal relations, we need to generate the proposal features as the input to the Proposal Transformer.
To do so, the previous works~\cite{lin2019bmn,su2020bsn++} usually utilize dense sampling, which will generate a lot of redundant proposals and cause an imbalance data distribution between positive/negative proposals and temporal durations.
Therefore, we propose a sparse sampling mechanism to generate the candidate proposals, then feed them into the Proposal Transformer which encodes the potential relationship between proposals, then output the corresponding confidence score from proposal confidence sequence.
We take the boundary probabilities and proposal confidences as input to \emph{Fuzzy matching} mechanism then get the matching proposals.
Followed by the post processing, we obtain the final proposal set.
We will describe the details of each module in the following sections.



\subsection{Boundary Transformer}
\label{sec:B-TR}
Long-term temporal dependency modeling is essential for boundary regression.
Previous methods usually conduct stacked temporal convolutions~\cite{lin2018bsn,lin2019bmn,lin2020fast} or global temporal pooling~\cite{gao2020accurate} to capture temporal relationship.
However, 1D temporal convolution can effectively capture local temporal relation between frames but lacks the capacity of modeling long-term temporal structure limited by the kernel size. On the contrary, the global temporal pooling performs average pooling to capture the global features over frames but fails to capture finer temporal structure and may introduce unnecessary noise.
To this end, we propose the Boundary Transformer to model both local temporal relation and long-range temporal dependency structure with a Transformer structure. 
The architecture of the Transformer is the same as the original Transformer model~\cite{vaswani2017attention}, which has been described in Sec.~\ref{sec:transformer}.
We adopt a $M$-layer Transformer and each layer has an encoder-decoder structure as follows:

The \textbf{Encoder} is applied to estimate the relevance of one frame to other frames.
First, we obtain the frame-wise feature sequence $F \in \mathbb{R}^{T \times C}$ using two-stream model~\cite{two-stream}, where $T$ is the length of the feature sequence and $C$ is the feature dimension.
Then we feed them to the Transformer encoder and generate an augmented feature sequence $F_g \in \mathbb{R}^{T \times C}$ with the global context.
The encoder has a standard architecture, which consists of a multi-head self-attention module and a feed forward network (FFN).
The outputs $F_g$ of the encoder along with the embeddings are given as input to the Transformer decoder.

The \textbf{Decoder} contains three sub-layers, first two (multi-head self-attention and feed forward network) are similar to the encoder, while the third sub-layer performs multi-head attention over the output of the encoder stack.
The input to the first self-attention module is the feature sequence $F \in \mathbb{R} ^{T \times C}$, which is the same as that of the encoder. Then the module output the feature sequence $F_q$.
In particular, for the second self-attention module, the $K$, $V$ are transformed from $F_g$ and $Q$ is transformed from $F_q$.
In this way, the decoder further enhances relevant features while reducing the over-smoothing effect from the decoder.

The output representation of decoder is then used as the global representation for the boundary prediction task. Specifically, we append a \textbf{Boundary Head} which encode the output representation of decoder with multi-layer perceptron (MLP) network and followed by a \emph{Sigmoid} layer to generate the predicted boundary probability sequence.

\subsection{Proposal Transformer}
\label{sec:P-TR}

For proposal confidence prediction, previous works~\cite{lin2019bmn} first generate densely distributed proposals then predict the confidence scores of them.
Obviously, dense proposals will result in imbalanced data distribution between positive/negative proposals and more computational burdens.
Meanwhile, the inter-proposal relationships modeling is a previously overlooked point, which plays a crucial role in proposal confidence evaluation.
To relieve these issues, we propose a sparse sampling mechanism and a Proposal Transformer as follows. 

The \textbf{Sparse sampling mechanism} aims to maintain the sparse proposal sequence.
Specifically, we use the Fibonacci sequence as a sliding window group $W=\{w_i\}_{i=1}^{D}$ to generate the proposals with different scales, where $D$ is the group size.
The step size of each sliding window is computed as:
\begin{equation}
step = \lfloor\sqrt{w_i} \rfloor + \lfloor \frac{w_i}{\gamma} \rfloor ,
\end{equation}
where $\gamma$ is used to constraint the step size increases with the increase of the window size.
For each proposal, we follow the BMN~\cite{lin2019bmn} and construct weight term $w_{i,j} \in \mathbb{R}^{N \times T}$ via uniformly sampling $N$ points between the temporal region. 
Then we conduct dot product in temporal dimension between $w_{i,j} \in \mathbb{R}^{N \times T}$ and $F \in \mathbb{R}^{T \times C}$ to generate the proposal feature with the shape ${C \times N}$.
Finally, we get the sparse proposal feature sequence $P \in \mathbb{R}^{L \times S}$ and these proposal features are subsequently flattened, where $L$ is number of proposal and $S = C \times N$. 

Moreover, we propose the Proposal Transformer to enhance the proposal feature representations for proposal confidence prediction.
We adopt a $M$-layer Transformer, which is based upon the encoder and the decoder in original Transformer model~\cite{vaswani2017attention}.
The \textbf{Encoder} aims to capture the relationships between proposals in various scales.
The sparse proposal feature sequence $P \in \mathbb{R}^{L \times S}$ is given as input to the encoder. 
Then the encoder outputs the feature sequence $P_g$ with the global view.
Similar to the Boundary Transformer, the decoder takes inputs from the encoder as well as the previous outputs to generate enhanced proposal feature sequence.
Taking the proposal sequence from the decoder output, our \textbf{Proposal Confidence Generator} will generate two types of confidence $C_c $, $ C_r$ $\in \mathbb{R}^{L \times 1}$ with \emph{Sigmoid} activation, where $C_c $, $ C_r$ are used for binary classification and regression loss function separately.


\subsection{Training}
\label{sec:train}
The overall objective of our framework is defined as:
\begin{equation}
    \mathcal{L} = \mathcal{L}_b + \mathcal{L}_p,
\end{equation}
where $\mathcal{L}_b$ and $\mathcal{L}_p$ are the objective functions of the Boundary Transformer and the Proposal Transformer respectively.

\textbf{Objective of Boundary Transformer.}
The Boundary Transformer generates the starting and ending probability sequence $P_s$, $P_e$.
We can define training objective of the Boundary Transformer as a two-task loss function. The loss function consists of starting loss and ending loss:
\begin{equation}
    \mathcal{L}_{b} = \mathcal{L}_{bl}(P_s, G_s) + \mathcal{L}_{bl}(P_e, G_e),
\end{equation}
where $G_s$ and $G_e$ are the ground truth labels of boundary sequence, and $\mathcal{L}_{bl}$ is the binary logistic regression loss function.

\textbf{Objective of Proposal Transformer.}
The Proposal Transformer generates sparse proposal confidence sequence $C_c$ and $C_r$.
The loss function $\mathcal{L}_{p}$ consists of binary classification loss and regression loss:
\begin{equation}
    \mathcal{L}_{p} = \mathcal{L}_c(C_c,G_c) + \mathcal{L}_r(C_r,G_c),
\end{equation}
where $\mathcal{L}_c$ is a binary logistic regression loss function, $ L_r$ is a smooth $L_1$ loss, and $G_c$ is the ground truth label of sparse proposal sequence.
The details of label assignment will be described in the supplementary material.

\subsection{Inference}
\label{sec:infer}
As mentioned above, the Boundary Transformer generates boundary probability sequence and the Proposal Transformer generates the sparse proposal confidence sequence.
Then we construct a coarse proposals set $\psi^c_p$ based on boundary probabilities.
Take the proposal $\varphi = [t_s',t_e'] \in {\psi}^c_p$ as an example, 
\textbf{Fuzzy matching} is devised to compute tIoU (temporal Intersection over Union) between proposal $\varphi$ and sparse proposal sequence, then select matching proposal $p_m = [t^m_s, t^m_e]$.
Next, we refine the proposal as:
\begin{equation}
   [t_s,t_e]=
    \begin{cases}
         [\frac{ t_s' + t^m_s}{2} , \frac{ t_e' + t^m_e}{2}], & if ~ {p_{m}^{cc} > {\alpha}_1} ~ and ~ {p_{m}^{cr} > {\alpha}_2} \\
         [t_s' ,  t_e'], &others
    \end{cases},
\end{equation}
where $p_{m}^{cc}$ is the proposal classification score, $p_{m}^{cr}$ is the regression confidence, ${\alpha}_1$ and ${\alpha}_2$ are the adjustment thresholds.
Finally, we get a proposal set $\psi_p = \{{\phi}_n = (t_s,t_e,p_{t_s'}^{s},p_{t_e'}^e,p_{m}^{cc},p_{m}^{cr})\}_{n=1}^N$, where $p_{t_s'}^{s},p_{t_e'}^{e}$ are the starting and ending probabilities.

Following the previous practices~\cite{lin2019bmn,lin2018bsn},
we also perform score fusion and redundant proposal suppression to further obtain final results.
Specifically, in order to make full use of various predicted scores for each proposal $\varphi_n$, we fuse its boundary probabilities and confidence scores of matching proposal by multiplication.
The confidence score $p^f$ can be defined as :
\begin{equation}
    p^f = p_{t_s'}^s \cdot p_{t_e'}^s \cdot p_{m}^{cc} \cdot  p_{m}^{cr}.
\end{equation}
Hence, the final proposal set as 
\begin{equation}
    \psi = {\{\varphi_n = (t_s,t_e,p^f)\}}_{n=1}^N.
\end{equation}
Moreover, we use the Soft-NMS~\cite{bodla2017soft} algorithm for \textbf{Post-processing} to remove the proposals which highly overlap with each other.


\section{Experiments}

\subsection{Datasets and Evaluation Metrics}
We evaluate our method on two challenge benchmarks: \textbf{ActivityNet-v1.3}~\cite{activitynet} and \textbf{THUMOS14}~\cite{thumos}.
ActivityNet-v1.3 is a large-scale video dataset for action recognition and temporal action detection tasks.
It contains 10K training, 5k validation, and 5k testing videos with 200 action categories, and the ratio of training, validation and testing sets is 2:1:1. Our models trained on the training set and evaluated with the validation set.
For THUMOS14, we use the subset of THUMOS14 that provides frame-wise action annotations. The model is trained with 200 untrimmed videos from its validation set and evaluated using 213 untrimmed videos from the test set. 
This dataset is challenging due to the large variations of the frequency and duration of action instances across videos.\\
\textbf{Evaluation Metrics. } 
Temporal action proposal generation aims to produce high quality proposals that have high IoU with ground truth action instances and high recall.
To evaluate quality of proposals, Average Recall (AR) is the average recall rate under specified tIoU thresholds for measuring the quality of proposals.
ActivityNetv1.3, these thresholds are set to [0.5:0.05:0.95]. 
On THUMOS14, they are set to [0.5:0.05:1.0].
By limiting the average number (AN) of proposals for each video , we can calculate the area under the AR \emph{vs.} AN curve to obtain AUC. 
We use \textbf{AR@AN} and \textbf{AUC} as our metrics to evaluate TAPG models.
For the temporal action detection task, mean Average Precision (\textbf{mAP}) under multiple tIoU is the widely used evaluation metric.
On ActivityNet-v1.3, the tIoU thresholds are set to \{0.5, 0.75, 0.95\}, and we also test the average mAP of tIoU thresholds between 0.5 and 0.95 with step of 0.05.
On THUMOS14, these tIoU thresholds are set to \{0.3, 0.4, 0.5, 0.6, 0.7 \}.

\subsection{Implementation Details.}
For feature extractor, following previous works~\cite{lin2019bmn,xu2020g}, we adopt the two-stream network~\cite{two-stream}, where ResNet~\cite{resnet} and BN-Inception network~\cite{bn} are used as the spatial and temporal networks respectively.
During feature extraction, the interval $\sigma$ is set to 16 and 5 on ActivityNet-1.3 and THUMOS14 respectively.
For ActivityNet-v1.3, each video feature sequence is rescaled to $T = 100$ snippets using linear interpolation.
For THUMOS14, we crop each video feature sequence with overlapped windows of size $T = 256$ and stride 128.
Besides, for the sparse sampling mechanism, the length of sparse proposal sequence $L$ is 451 and 1225 on ActivityNet-1.3 and THUMOS14, the max sliding window size $D$ is 100 and 64 for ActivityNet-v1.3 and THUMOS14, respectively, and the constraint hyper-parameter $\gamma$ is 21.
For each proposal, we sampling points $N = 32$.
For Fuzzy matching, we set adjustment thresholds ${\alpha}_1 = 0.9$  and ${\alpha}_2 =0.8 $.
Due to the limit of computation resource, we apply 1D Conv for dimension reduction, then take the features as the input to the Boundary Transformer and Proposal Transformer.
For temporal action detection task, following~\cite{xu2020g}, we take the video classification scores from~\cite{xiong2016cuhk} and~\cite{wang2017untrimmednets} and multiply them to yield the fused confidence score $p^f$ for evaluation.
We train our TAPG Transformer from scratch using the Adam optimizer and the learning rate is set to $10^{-4}$ and decayed by a factor of 0.1 after every 10 epoch.
To preserve the positional information in the transformer, we adopt sine positional encoding for queries and keys.
The number of Transformer layers $M=3$.

\subsection{Comparison with State-of-the-art Results}
In this subsection, we compare our method with the existing state-of-the-art methods on ActivityNet-v1.3 and THUMOS14.
For a fair comparison, we adopt the same two-stream features used by previous methods in our experiments.

\begin{table}[th]
\caption{Performance comparison with state-of-the-art proposal generation methods on test set of THUMOS14 in terms of AR@AN.}
\centering
\scalebox{0.95}{
\begin{tabular}{cccccc}
\hline

  Method &@50 &@100 &@200 &@500 &@1000 \\ \hline
 TAG~\cite{zhao2017temporal}&18.6 & 29.0 & 39.6 & - & - \\
 CTAP~\cite{gao2018ctap} &32.5 & 42.6 & 52.0 & - & - \\
 BSN~\cite{lin2018bsn}& 37.5 & 46.1 & 53.2 & 61.4 & 65.1 \\
 MGG~\cite{liu2019multi}&39.9 & 47.8 & 54.7 &61.4 & 64.6 \\
 BMN~\cite{lin2019bmn}&39.4 & 47.7 & 54.8 & 62.2 & 65.5 \\
 BSN++ ~\cite{su2020bsn++}&42.4 & 49.8 & 57.6 & 65.2 & 66.8 \\
 TCANet~\cite{qing2021temporal}&42.1 & 50.5 & 57.1 & 63.61 & 66.9 \\
  RTD-Net~\cite{tan2021relaxed}& 41.1 & 49.0 &56.1 & 62.9 & - \\ \hline
\textbf{Ours} & \textbf{43.9} & \textbf{53.5} & \textbf{60.2} & \textbf{66.7} & \textbf{68.8} \\

\hline
\end{tabular}
}

\label{tab:thumos_ar_an}
\end{table}

\begin{table}[th]
\caption{Performance comparison with state-of-the-art action detection methods on test set of THUMOS14, in terms of mAP (\%) at different tIoU thresholds.}
\centering
\scalebox{0.95}{
\begin{tabular}{cccccc}
\hline

Method  &0.3 &0.4 &0.5 &0.6&0.7 \\ \hline
SST~\cite{kaustend}& - & - &23.0 &- &- \\
CDC~\cite{shou2017cdc}& 40.1 &  &23.0 &- &- \\
TURN-TAP~\cite{gao2017turn}& 44.1 & 34.9 & 25.6 & - &- \\
SSN~\cite{yue2015beyond}& 51.9 & 41.0 &29.8 &- &- \\
BSN~\cite{lin2018bsn}& 53.5 & 45.0 &36.9 &28.4 &20.0 \\
TAL-Net~\cite{chao2018rethinking}& 53.2 & 48.5 &42.8 &33.8 &20.8 \\
MGG~\cite{liu2019multi}& 53.9 & 46.8 &37.4 &29.5 &21.3 \\
DBG~\cite{lin2020fast}& 57.8 & 49.4 &39.8 &30.2 &21.7 \\
BMN~\cite{lin2019bmn}& 56.0 & 47.4 &38.8 &29.7 &20.5 \\
G-TAD\cite{xu2020g}& 54.5 & 47.6 &40.2 &30.8 &23.4 \\
BSN++~\cite{su2020bsn++}& 59.9 & 49.5 &41.3 &31.9 &22.8 \\
TCANet~\cite{qing2021temporal}& 60.6 & 53.2 &44.6 &36.8 &26.7 \\
RTD-Net~\cite{tan2021relaxed}& 53.9 & 48.9 & 42.0 & 33.9 & 23.4 \\ \hline
\textbf{Ours} & \textbf{65.3} & \textbf{59.3} & \textbf{50.8} & \textbf{40.1} & \textbf{28.5} \\

\hline
\end{tabular}
}

\label{tab:thumos_map}
\end{table}

\textbf{Results on THUMOS14.} 
We present performance comparison of the proposed TAPG Transformer with the state-of-the-art methods on THUMOS14 in Table~\ref{tab:thumos_ar_an} and Table~\ref{tab:thumos_map}, where our method improves the performance significantly for both temporal action proposal generation and action detection.
For the temporal action proposal generation task, results are shown in Table~\ref{tab:thumos_ar_an}, which demonstrate that TAPG Transformer outperforms state-of-the-art methods in terms of AR@AN with AN varying from 50 to 1000.
For the temporal action detection task, the proposed TAPG Transformer also achieves superior results, as shown in Table~\ref{tab:thumos_map}. The performance of our method exceeds state-of-the-art proposal generation methods by a big margin at different tIoU thresholds. 
The high mAP reflects that our model is able to predict the proposals with high scores, and reduce the number of false positives.

\begin{table}[th]
\caption{Performance comparison with state-of-the-art proposal generation methods on validation set of ActivityNet-1.3 in terms of AUC and AR@AN.}
\centering
\scalebox{0.95}{
\begin{tabular}{cccc}
\hline
Method & AR@1 (val) & AR@100 (val) &AUC (val) \\ \hline
CTAP~\cite{gao2018ctap} & - & 73.2  & 65.7 \\
BSN~\cite{lin2018bsn} & 32.2 & 74.2 & 66.2 \\
MGG~\cite{liu2019multi} & - & 75.5 & 66.4 \\
BMN~\cite{lin2019bmn}& - & 75.0 & 67.0 \\
BSN++~\cite{su2020bsn++}& 34.3 & 76.5 & 68.3 \\
TCANet ~\cite{qing2021temporal}& 34.6 & 76.1 & 68.1 \\
RTD-Net~\cite{tan2021relaxed}& 32.8 & 73.1&65.7\\ \hline
\textbf{Ours} & \textbf{34.9} & \textbf{76.5} & \textbf{68.3} \\
\hline
\end{tabular}
}
\label{tab:anet_ar_an}
\end{table}

\begin{table*}[th]
		\caption{{Ablation study} on \textbf{THUMOS14}. We show mAP (\%) at different tIoU thresholds.
		}
        \begin{subtable}[th]{0.44\textwidth}
        \centering
        \caption{Study on different combinations of the components in TAPG Transformer and BMN.
        BTR denotes Boundary Transformer and PTR denotes Proposal Transformer.
        }
        \scalebox{0.95}{
        \begin{tabular}{ccccccc}
        \hline
        Model & Module  &0.3 &0.4 &0.5 &0.6&0.7 \\ \hline
        BMN & TEM & 41.9 &36.4 & 27.8 & 21.6& 12.7 \\
        BMN & TEM+PEM & 56.0 & 47.4 & 38.8 & 29.7& 20.5 \\ \hline
        Ours & BTR & 43.8 & 37.7 & 29.9 & 20.8& 12.9\\
        Ours & BTR + PEM& 60.6 & 54.5 & 46.3 & 37.6& 27.1\\
        Ours & TEM + PTR & 62.9 & 57.6 & 47.6 & 37.5& 25.3\\
        Ours & BTR + PTR & \textbf{65.3} & \textbf{59.3} & \textbf{50.8} & \textbf{40.1}& \textbf{28.5}\\
        \hline
        \end{tabular}
        }
        \label{tab:ablation module}
        \end{subtable}		
	    \hspace{4mm}
        \begin{subtable}[th]{0.52\textwidth}
        \caption{Study on sparse sampling mechanism and different window groups.
        [\emph{start}:\emph{stride}:\emph{end}] means that the sliding window size varies from \emph{start} to \emph{end} with \emph{stride}.
        }        
        \scalebox{0.95}{
        \begin{tabular}{cccccccc}
        \hline
        
        Model & Number & window group  &0.3 &0.4 &0.5 &0.6&0.7 \\ \hline
		BMN &16384 &Densely &56.0 & 47.4  & 38.8  & 29.7 & 20.5\\
        BMN &1225&Fibonacci &58.2&50.5  & 41.2  & 32.9 & 23.3\\ \hline
		Ours &800 &  [2:5:62]   & 63.5 &57.6  & 49.8  & 40.5 & \textbf{29.4} \\ 
        Ours &1218 &  [2:3:62]   & 64.6 & 58.5  & 50.0  & 40.4 & 28.1 \\ 
        Ours &1726 &  [2:2:64]   & 65.2 & 59.1  & 50.8  & \textbf{41.0 }& 28.8 \\ 
        Ours &1225 & Fibonacci  & \textbf{65.3} & \textbf{59.3}  & \textbf{50.8}  & 40.1 & 28.5 \\ 
        \hline
        \end{tabular}
        }
        \label{tab:ablation candidate proposal number}
        \end{subtable}
		\\[7pt]
        \begin{subtable}[th]{0.48\textwidth}
        \centering
        \caption{Study on different query forms of transformer decoder.}        
        \scalebox{0.95}{
        \begin{tabular}{cccccc}
        \hline
        Query  &0.3 &0.4 &0.5 &0.6&0.7 \\ \hline
		Learnable parameters & 62.1 & 54.1 & 45.8 & 34.8 & 23.2 \\
		Encoder output & 64.2 & 57.4 & 48.7 & 39.1 & 26.8 \\
		Original Feature & \textbf{65.3} & \textbf{59.3} & \textbf{50.8} & \textbf{40.1 }&\textbf{ 28.5} \\
        \hline
        \end{tabular}
        }
        \label{tab:ablation_query}
        \end{subtable}
		\hspace{2mm}
        \begin{subtable}[th]{0.48\textwidth}
        \caption{Study on different numbers of TAPG Transformer layers.}
        \centering
        \scalebox{0.95}{
        \begin{tabular}{cccccc}
        \hline
        M  &0.3 &0.4 &0.5 &0.6&0.7 \\ \hline
        1 & 64.8 & 59.0 & 50.5 & 40.0& 27.9 \\
        3 & \textbf{65.3} & \textbf{59.3} & \textbf{50.8} & \textbf{40.1}& \textbf{28.5} \\
        6 & 64.2 & 57.6 & 49.2 & 38.9& 26.4 \\
        \hline
        \end{tabular}
        }
        \label{tab:ablation_layers}
        \end{subtable}	
		\label{tab:ablations}
	\end{table*}
\begin{table}
\caption{Performance comparison with state-of-the-art action detection methods on validation set of ActivityNet-1.3, in terms of mAP (\%) at different tIoU thresholds and the average mAP.}
\centering
\scalebox{0.95}{
\begin{tabular}{ccccc}
\hline

Method  &0.5 &0.75 &0.95 &Average \\ \hline
Singh et al.~\cite{singh2016untrimmed}& 34.5 &- &- &- \\
SCC~\cite{heilbron2017scc}& 40.0 & 17.9 & 4.7 & 21.7 \\
CDC~\cite{shou2017cdc}& 45.3 & 26.0 & 0.20 &23.8 \\
R-C3D~\cite{xu2017r}& 26.8 &- &- &- \\
BSN~\cite{lin2018bsn}& 46.5& 30.0 & 8.0 & 30.0 \\
BMN~\cite{lin2019bmn}& 50.1 & 34.8 & 8.3 & 33.9 \\
GTAD~\cite{xu2020g}& 50.4 & 34.6 & 9.0 & 35.1 \\
BSN++~\cite{su2020bsn++}& 51.3 & \textbf{35.7} & 8.3 & 34.9 \\
TCANet \emph{w/} BSN~\cite{qing2021temporal}& 51.9 & 34.9 & 7.5 & 34.4 \\
RTD-Net~\cite{tan2021relaxed}&46.4 & 30.5 & 8.6 & 30.5 \\ \hline
\textbf{Ours} & \textbf{53.2} & 35.5 & \textbf{10.1} & \textbf{35.4}\\

\hline
\end{tabular}
}

\label{tab:anet_map}
\end{table}

\textbf{Results on ActivityNet-v1.3.} 
Table ~\ref{tab:anet_ar_an} and Table~\ref{tab:anet_map} shows comparison results of the proposed TAPG Transformer with other methods on ActivityNet-v1.3.
For the temporal action proposal generation task, as shown in Table~\ref{tab:anet_ar_an}, the performance of TAPG Transformer again outperforms state-of-the-art proposal generation methods in terms of AR@AN with AN varying from 1 to 100 and AUC.
Especially when AN equals 1, we achieve 34.9\% regarding the AR metric, which indicates that top-1 proposal has high quality.
For the temporal action detection task, as summarized in Table~\ref{tab:anet_map},
our method achieves notable improvements on mAP over other proposal generation methods such as BSN~\cite{lin2018bsn}, BMN~\cite{lin2019bmn} and G-TAD~\cite{xu2020g} at all tIoU thresholds.
When tIoU is 0.95, the mAP we obtain is 10.1\%, indicating that the boundaries of the generated proposals by TAPG Transformer are more precise.



\subsection{Ablation Study}
In this section, we conduct ablation studies on THUMOS14 to verify the effectiveness of each module in TAPG Transformer.

\textbf{Effectiveness of Transformers.}
We perform ablation studies under different architecture settings and verify the effectiveness of the Boundary Transformer (BTR) and Proposal Transformer (PTR), separately.
Table~\ref{tab:ablation module} demonstrates that both modules improve the performance individually. 
Temporal Evaluation Module (TEM) and Proposal Evaluation Module (PEM) are two modules of BMN \cite{lin2019bmn}.
As can be seen, "BTR + PEM" outperforming "TEM + PEM" indicates that BTR achieves more accurate boundaries, as shown in the second and fourth rows of Table~\ref{tab:ablation module}, while "TEM + PTR" outperforming "TEM + PEM" demonstrates that PTR improves the proposal confidence, as shown in the second and fifth rows of Table~\ref{tab:ablation module}. 
To summarize, the evaluation results shown in Table~\ref{tab:ablation module} demonstrate that:
1) Compared with the previous works which only explore the local details, Boundary Transformer effectively captures the long-term information for accurate boundary prediction. 
2) Instead of treating the proposals separately, Proposal Transformer effectively learns the relation of the proposals on different scales to generate high quality proposal confidence.

\textbf{Effectiveness of Sparse Sampling mechanism.}
We perform ablation studies to verify the effectiveness of sparse proposal sequence and different sampling methods.
Generating dense BM map to predict proposal confidence as done in~\cite{lin2019bmn} causes  unbalanced positive/negative proposals and resource waste.
As shown in the first two rows of Table~\ref{tab:ablation candidate proposal number}, generating less proposals (about only \textbf{7.5\%} of the BM maps) still achieves performance improvement.
Meanwhile we design two ways to generate window group.
The first is using different window sizes with equal stride in each group. The third to fifth rows in Table~\ref{tab:ablation candidate proposal number} shows that generating sliding windows with the stride set to 5, 3, and 2 all obtain good results.
The second is using different window sizes with different distances in each group, specifically, we apply the Fibonacci sequence to generate candidate proposals. Satisfying performance is achieved as shown in the last row of Table~\ref{tab:ablation candidate proposal number}.
In summary, the proposed TAPG Transformer achieves better performance with fewer proposals.
\begin{figure*}[t]
	\begin{center}
	\includegraphics[width=1\textwidth]{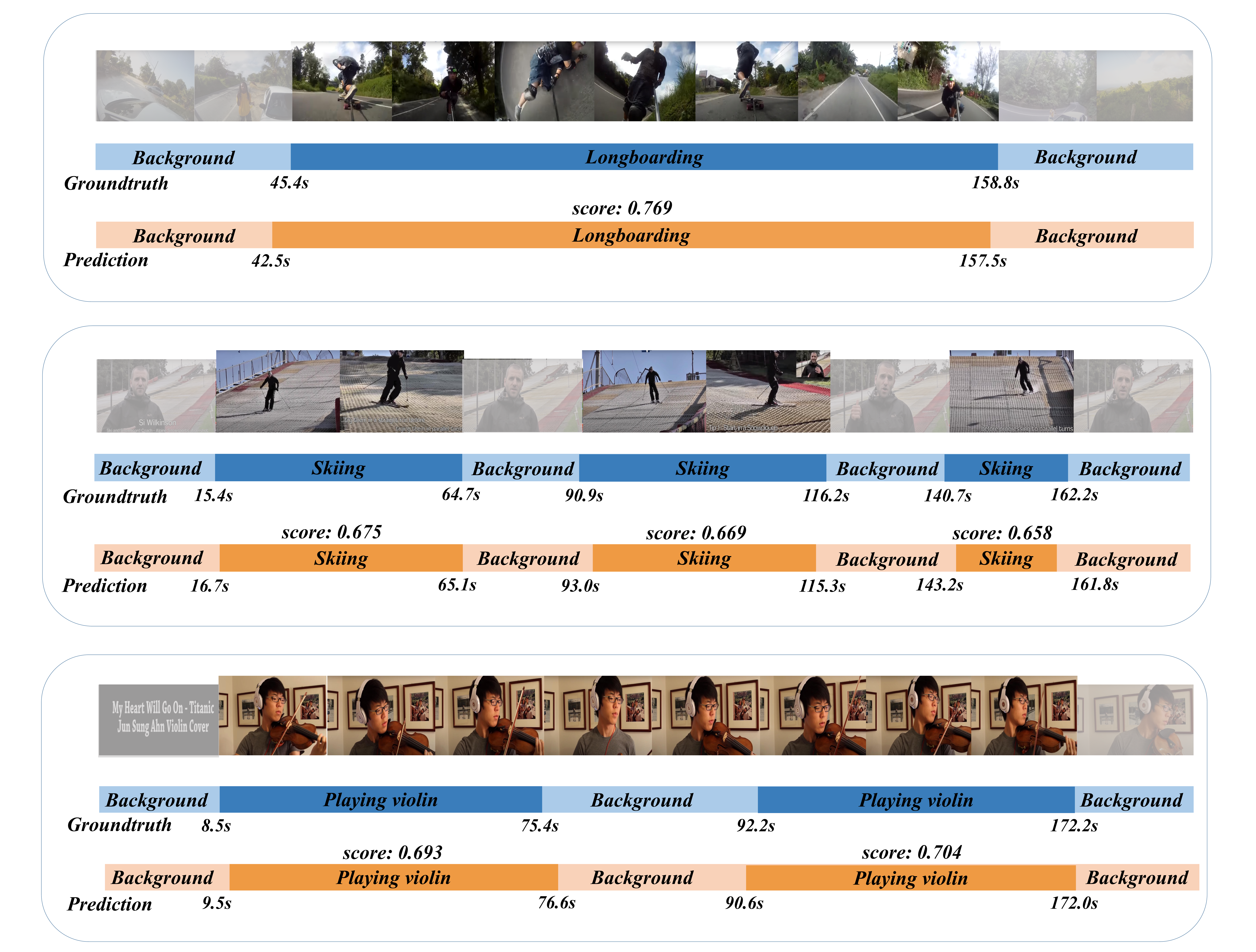}
	\end{center}
	\caption{Visualization examples of generated proposals on ActivityNet1.3.}
   	\label{fig:example}
\end{figure*}

\textbf{Effectiveness of Query Form of Transformer Decoder.}
We perform ablation studies to explore the performance of Transformer Decoder with different query forms.
Long-term modeling easily leads to the smooth boundary of action and introduces global noise.
To solve this problem, we design novel query forms of transformer decoder that are different from original Transformer.
Specifically, we study on three query forms including learnable parameters~\cite{carion2020end}, encoder output and original feature. 
The experimental results shown in Table~\ref{tab:ablation_query} demonstrate that the query form of original feature achieves the best performance.
We believe the reason is that introducing original feature helps eliminate the effects of smooth boundary of action and the global noise. 

\textbf{Effectiveness of Layers of Transformer.}
Stacking multi-layer transformers normally brings better performance, which has been proved in multiple tasks, \emph{e.g.}, NLP~\cite{vaswani2017attention} and object detection~\cite{carion2020end}.
We study on the different number of the stack of multiple layers in Boundary Transformer and Proposal Transformer.
As shown in Table~\ref{tab:ablation_layers}, we achieve the best performance when stacking three Transformer modules.
However, stacking an increasing number of layers does not necessarily lead to performance improvement, and the stack of excessive layers may lead to over-fitting as well, as shown in the last row of Table~\ref{tab:ablation_layers}.

\subsection{Visualization}
Figure ~\ref{fig:example} presents visualization examples of some representative and challenging cases on ActivityNet1.3.
The proposals with the highest $k$ scores are visualized in each video, where $k$ is the number of ground truth.
In the top video, although there exist scenes of falling and the first-person perspective, our method predicts the video correctly as a continuous action. 
The correct prediction shows that our method understands the inter-frame relation and inter-proposal relation properly and does not overfit the noisy frames.
The middle video includes multiple ground truth action instances, while our top-3 proposals match with them accurately, indicating the high quality of our generated proposals.
In the bottom video, where the background frames and action frames are of a similar scene, our top-2 proposals are successfully aligned with the positions of ground truth action instances.
Our method considers the high-level semantic information and can identify the noisy snippets even when they have a very similar appearance to the action.
More examples and comparison results with the previous methods are shown in the supplementary material.

\section{Conclusion}
In this paper, we propose a novel temporal action proposal generation framework with transformers, which consists a Boundary Transformer and a Proposal Transformer.
Boundary Transformer captures long-term temporal context and Proposal Transformer models inter-proposal relation.
Our TAPG Transformer can effectively enhance video understanding and noisy action instance localization.
Extensive experiments show that our model achieves new state-of-the-art performance in temporal action proposal generation and action detection on THUMOS14 and ActivityNet1.3 datasets.
\bibliographystyle{ACM-Reference-Format}
\bibliography{sample-base}


\begin{thebibliography}{58}


\ifx \showCODEN    \undefined \def \showCODEN     #1{\unskip}     \fi
\ifx \showDOI      \undefined \def \showDOI       #1{#1}\fi
\ifx \showISBNx    \undefined \def \showISBNx     #1{\unskip}     \fi
\ifx \showISBNxiii \undefined \def \showISBNxiii  #1{\unskip}     \fi
\ifx \showISSN     \undefined \def \showISSN      #1{\unskip}     \fi
\ifx \showLCCN     \undefined \def \showLCCN      #1{\unskip}     \fi
\ifx \shownote     \undefined \def \shownote      #1{#1}          \fi
\ifx \showarticletitle \undefined \def \showarticletitle #1{#1}   \fi
\ifx \showURL      \undefined \def \showURL       {\relax}        \fi
\providecommand\bibfield[2]{#2}
\providecommand\bibinfo[2]{#2}
\providecommand\natexlab[1]{#1}
\providecommand\showeprint[2][]{arXiv:#2}

\bibitem[\protect\citeauthoryear{Bertasius, Wang, and Torresani}{Bertasius
  et~al\mbox{.}}{2021}]%
        {bertasius2021space}
\bibfield{author}{\bibinfo{person}{Gedas Bertasius}, \bibinfo{person}{Heng
  Wang}, {and} \bibinfo{person}{Lorenzo Torresani}.}
  \bibinfo{year}{2021}\natexlab{}.
\newblock \showarticletitle{Is Space-Time Attention All You Need for Video
  Understanding?}
\newblock \bibinfo{journal}{\emph{arXiv preprint arXiv:2102.05095}}
  (\bibinfo{year}{2021}).
\newblock


\bibitem[\protect\citeauthoryear{Bodla, Singh, Chellappa, and Davis}{Bodla
  et~al\mbox{.}}{2017}]%
        {bodla2017soft}
\bibfield{author}{\bibinfo{person}{Navaneeth Bodla}, \bibinfo{person}{Bharat
  Singh}, \bibinfo{person}{Rama Chellappa}, {and} \bibinfo{person}{Larry~S
  Davis}.} \bibinfo{year}{2017}\natexlab{}.
\newblock \showarticletitle{Soft-NMS--improving object detection with one line
  of code}. In \bibinfo{booktitle}{\emph{Proceedings of the IEEE international
  conference on computer vision}}. \bibinfo{pages}{5561--5569}.
\newblock


\bibitem[\protect\citeauthoryear{Buch, Escorcia, Ghanem, Fei-Fei, and
  Niebles}{Buch et~al\mbox{.}}{2017}]%
        {kaustend}
\bibfield{author}{\bibinfo{person}{Shyamal Buch}, \bibinfo{person}{Victor
  Escorcia}, \bibinfo{person}{Bernard Ghanem}, \bibinfo{person}{Li Fei-Fei},
  {and} \bibinfo{person}{Juan~Carlos Niebles}.}
  \bibinfo{year}{2017}\natexlab{}.
\newblock \showarticletitle{End-to-End, Single-Stream Temporal Action Detection
  in Untrimmed Videos}. In \bibinfo{booktitle}{\emph{BMVC 2017}}.
\newblock


\bibitem[\protect\citeauthoryear{Caba~Heilbron, Escorcia, Ghanem, and
  Carlos~Niebles}{Caba~Heilbron et~al\mbox{.}}{2015}]%
        {activitynet}
\bibfield{author}{\bibinfo{person}{Fabian Caba~Heilbron},
  \bibinfo{person}{Victor Escorcia}, \bibinfo{person}{Bernard Ghanem}, {and}
  \bibinfo{person}{Juan Carlos~Niebles}.} \bibinfo{year}{2015}\natexlab{}.
\newblock \showarticletitle{Activitynet: A large-scale video benchmark for
  human activity understanding}. In \bibinfo{booktitle}{\emph{Proceedings of
  the ieee conference on computer vision and pattern recognition}}.
  \bibinfo{pages}{961--970}.
\newblock


\bibitem[\protect\citeauthoryear{Carion, Massa, Synnaeve, Usunier, Kirillov,
  and Zagoruyko}{Carion et~al\mbox{.}}{2020}]%
        {carion2020end}
\bibfield{author}{\bibinfo{person}{Nicolas Carion}, \bibinfo{person}{Francisco
  Massa}, \bibinfo{person}{Gabriel Synnaeve}, \bibinfo{person}{Nicolas
  Usunier}, \bibinfo{person}{Alexander Kirillov}, {and} \bibinfo{person}{Sergey
  Zagoruyko}.} \bibinfo{year}{2020}\natexlab{}.
\newblock \showarticletitle{End-to-end object detection with transformers}. In
  \bibinfo{booktitle}{\emph{European Conference on Computer Vision}}. Springer,
  \bibinfo{pages}{213--229}.
\newblock


\bibitem[\protect\citeauthoryear{Carreira and Zisserman}{Carreira and
  Zisserman}{2017}]%
        {i3d}
\bibfield{author}{\bibinfo{person}{Joao Carreira} {and} \bibinfo{person}{Andrew
  Zisserman}.} \bibinfo{year}{2017}\natexlab{}.
\newblock \showarticletitle{Quo Vadis, Action Recognition? A New Model and the
  Kinetics Dataset}. In \bibinfo{booktitle}{\emph{CVPR}}.
\newblock


\bibitem[\protect\citeauthoryear{Chao, Vijayanarasimhan, Seybold, Ross, Deng,
  and Sukthankar}{Chao et~al\mbox{.}}{2018}]%
        {chao2018rethinking}
\bibfield{author}{\bibinfo{person}{Yu-Wei Chao}, \bibinfo{person}{Sudheendra
  Vijayanarasimhan}, \bibinfo{person}{Bryan Seybold}, \bibinfo{person}{David~A
  Ross}, \bibinfo{person}{Jia Deng}, {and} \bibinfo{person}{Rahul Sukthankar}.}
  \bibinfo{year}{2018}\natexlab{}.
\newblock \showarticletitle{Rethinking the faster r-cnn architecture for
  temporal action localization}. In \bibinfo{booktitle}{\emph{Proceedings of
  the IEEE Conference on Computer Vision and Pattern Recognition}}.
  \bibinfo{pages}{1130--1139}.
\newblock


\bibitem[\protect\citeauthoryear{Chen, Yan, Zhu, Wang, Yang, and Lu}{Chen
  et~al\mbox{.}}{2021}]%
        {chen2021transformer}
\bibfield{author}{\bibinfo{person}{Xin Chen}, \bibinfo{person}{Bin Yan},
  \bibinfo{person}{Jiawen Zhu}, \bibinfo{person}{Dong Wang},
  \bibinfo{person}{Xiaoyun Yang}, {and} \bibinfo{person}{Huchuan Lu}.}
  \bibinfo{year}{2021}\natexlab{}.
\newblock \showarticletitle{Transformer Tracking}.
\newblock \bibinfo{journal}{\emph{arXiv preprint arXiv:2103.15436}}
  (\bibinfo{year}{2021}).
\newblock


\bibitem[\protect\citeauthoryear{Dosovitskiy, Beyer, Kolesnikov, Weissenborn,
  Zhai, Unterthiner, Dehghani, Minderer, Heigold, Gelly,
  et~al\mbox{.}}{Dosovitskiy et~al\mbox{.}}{2020}]%
        {dosovitskiy2020image}
\bibfield{author}{\bibinfo{person}{Alexey Dosovitskiy}, \bibinfo{person}{Lucas
  Beyer}, \bibinfo{person}{Alexander Kolesnikov}, \bibinfo{person}{Dirk
  Weissenborn}, \bibinfo{person}{Xiaohua Zhai}, \bibinfo{person}{Thomas
  Unterthiner}, \bibinfo{person}{Mostafa Dehghani}, \bibinfo{person}{Matthias
  Minderer}, \bibinfo{person}{Georg Heigold}, \bibinfo{person}{Sylvain Gelly},
  {et~al\mbox{.}}} \bibinfo{year}{2020}\natexlab{}.
\newblock \showarticletitle{An image is worth 16x16 words: Transformers for
  image recognition at scale}.
\newblock \bibinfo{journal}{\emph{arXiv preprint arXiv:2010.11929}}
  (\bibinfo{year}{2020}).
\newblock


\bibitem[\protect\citeauthoryear{Escorcia, Heilbron, Niebles, and
  Ghanem}{Escorcia et~al\mbox{.}}{2016}]%
        {escorcia2016daps}
\bibfield{author}{\bibinfo{person}{Victor Escorcia},
  \bibinfo{person}{Fabian~Caba Heilbron}, \bibinfo{person}{Juan~Carlos
  Niebles}, {and} \bibinfo{person}{Bernard Ghanem}.}
  \bibinfo{year}{2016}\natexlab{}.
\newblock \showarticletitle{Daps: Deep action proposals for action
  understanding}. In \bibinfo{booktitle}{\emph{European Conference on Computer
  Vision}}. Springer, \bibinfo{pages}{768--784}.
\newblock


\bibitem[\protect\citeauthoryear{Feichtenhofer}{Feichtenhofer}{2020}]%
        {feichtenhofer2020x3d}
\bibfield{author}{\bibinfo{person}{Christoph Feichtenhofer}.}
  \bibinfo{year}{2020}\natexlab{}.
\newblock \showarticletitle{X3D: Expanding Architectures for Efficient Video
  Recognition}. In \bibinfo{booktitle}{\emph{IEEE Conf. Comput. Vis. Pattern
  Recog.}} \bibinfo{pages}{203--213}.
\newblock


\bibitem[\protect\citeauthoryear{Feichtenhofer, Fan, Malik, and
  He}{Feichtenhofer et~al\mbox{.}}{2019}]%
        {slowfast}
\bibfield{author}{\bibinfo{person}{Christoph Feichtenhofer},
  \bibinfo{person}{Haoqi Fan}, \bibinfo{person}{Jitendra Malik}, {and}
  \bibinfo{person}{Kaiming He}.} \bibinfo{year}{2019}\natexlab{}.
\newblock \showarticletitle{Slowfast networks for video recognition}. In
  \bibinfo{booktitle}{\emph{Proceedings of the IEEE/CVF International
  Conference on Computer Vision}}. \bibinfo{pages}{6202--6211}.
\newblock


\bibitem[\protect\citeauthoryear{Gao, Chen, and Nevatia}{Gao
  et~al\mbox{.}}{2018}]%
        {gao2018ctap}
\bibfield{author}{\bibinfo{person}{Jiyang Gao}, \bibinfo{person}{Kan Chen},
  {and} \bibinfo{person}{Ram Nevatia}.} \bibinfo{year}{2018}\natexlab{}.
\newblock \showarticletitle{Ctap: Complementary temporal action proposal
  generation}. In \bibinfo{booktitle}{\emph{Proceedings of the European
  conference on computer vision (ECCV)}}. \bibinfo{pages}{68--83}.
\newblock


\bibitem[\protect\citeauthoryear{Gao, Shi, Li, Wang, Yuan, Ge, and Zhou}{Gao
  et~al\mbox{.}}{2020}]%
        {gao2020accurate}
\bibfield{author}{\bibinfo{person}{Jialin Gao}, \bibinfo{person}{Zhixiang Shi},
  \bibinfo{person}{Jiani Li}, \bibinfo{person}{Guanshuo Wang},
  \bibinfo{person}{Yufeng Yuan}, \bibinfo{person}{Shiming Ge}, {and}
  \bibinfo{person}{Xi Zhou}.} \bibinfo{year}{2020}\natexlab{}.
\newblock \showarticletitle{Accurate Temporal Action Proposal Generation with
  Relation-Aware Pyramid Network}.
\newblock \bibinfo{journal}{\emph{arXiv preprint arXiv:2003.04145}}
  (\bibinfo{year}{2020}).
\newblock


\bibitem[\protect\citeauthoryear{Gao, Yang, Sun, Chen, and Nevatia}{Gao
  et~al\mbox{.}}{2017}]%
        {gao2017turn}
\bibfield{author}{\bibinfo{person}{Jiyang Gao}, \bibinfo{person}{Zhenheng
  Yang}, \bibinfo{person}{Chen Sun}, \bibinfo{person}{Kan Chen}, {and}
  \bibinfo{person}{Ram Nevatia}.} \bibinfo{year}{2017}\natexlab{}.
\newblock \showarticletitle{Turn tap: Temporal unit regression network for
  temporal action proposals}. In \bibinfo{booktitle}{\emph{ICCV}}.
\newblock


\bibitem[\protect\citeauthoryear{He, Zhang, Ren, and Sun}{He
  et~al\mbox{.}}{2016}]%
        {resnet}
\bibfield{author}{\bibinfo{person}{Kaiming He}, \bibinfo{person}{Xiangyu
  Zhang}, \bibinfo{person}{Shaoqing Ren}, {and} \bibinfo{person}{Jian Sun}.}
  \bibinfo{year}{2016}\natexlab{}.
\newblock \showarticletitle{Deep residual learning for image recognition}. In
  \bibinfo{booktitle}{\emph{Proceedings of the IEEE conference on computer
  vision and pattern recognition}}. \bibinfo{pages}{770--778}.
\newblock


\bibitem[\protect\citeauthoryear{Heilbron, Barrios, Escorcia, and
  Ghanem}{Heilbron et~al\mbox{.}}{2017}]%
        {heilbron2017scc}
\bibfield{author}{\bibinfo{person}{F~Caba Heilbron}, \bibinfo{person}{Wayner
  Barrios}, \bibinfo{person}{Victor Escorcia}, {and} \bibinfo{person}{Bernard
  Ghanem}.} \bibinfo{year}{2017}\natexlab{}.
\newblock \showarticletitle{SCC: Semantic context cascade for efficient action
  detection}. In \bibinfo{booktitle}{\emph{CVPR}}.
\newblock


\bibitem[\protect\citeauthoryear{Ioffe and Szegedy}{Ioffe and Szegedy}{2015}]%
        {bn}
\bibfield{author}{\bibinfo{person}{Sergey Ioffe} {and}
  \bibinfo{person}{Christian Szegedy}.} \bibinfo{year}{2015}\natexlab{}.
\newblock \showarticletitle{Batch Normalization: Accelerating Deep Network
  Training by Reducing Internal Covariate Shift}.
\newblock


\bibitem[\protect\citeauthoryear{Jiang, Liu, Zamir, Toderici, Laptev, Shah, and
  Sukthankar}{Jiang et~al\mbox{.}}{2014}]%
        {thumos}
\bibfield{author}{\bibinfo{person}{Yu-Gang Jiang}, \bibinfo{person}{Jingen
  Liu}, \bibinfo{person}{A~Roshan Zamir}, \bibinfo{person}{George Toderici},
  \bibinfo{person}{Ivan Laptev}, \bibinfo{person}{Mubarak Shah}, {and}
  \bibinfo{person}{Rahul Sukthankar}.} \bibinfo{year}{2014}\natexlab{}.
\newblock \bibinfo{title}{THUMOS challenge: Action recognition with a large
  number of classes}.
\newblock
\newblock


\bibitem[\protect\citeauthoryear{Korbar, Tran, and Torresani}{Korbar
  et~al\mbox{.}}{2019}]%
        {korbar2019scsampler}
\bibfield{author}{\bibinfo{person}{Bruno Korbar}, \bibinfo{person}{Du Tran},
  {and} \bibinfo{person}{Lorenzo Torresani}.} \bibinfo{year}{2019}\natexlab{}.
\newblock \showarticletitle{Scsampler: Sampling salient clips from video for
  efficient action recognition}. In \bibinfo{booktitle}{\emph{Int. Conf.
  Comput. Vis.}} \bibinfo{pages}{6232--6242}.
\newblock


\bibitem[\protect\citeauthoryear{Kumar, Weissenborn, and Kalchbrenner}{Kumar
  et~al\mbox{.}}{2021}]%
        {kumar2021colorization}
\bibfield{author}{\bibinfo{person}{Manoj Kumar}, \bibinfo{person}{Dirk
  Weissenborn}, {and} \bibinfo{person}{Nal Kalchbrenner}.}
  \bibinfo{year}{2021}\natexlab{}.
\newblock \showarticletitle{Colorization transformer}.
\newblock \bibinfo{journal}{\emph{arXiv preprint arXiv:2102.04432}}
  (\bibinfo{year}{2021}).
\newblock


\bibitem[\protect\citeauthoryear{Li, Ji, Shi, Zhang, Kang, and Wang}{Li
  et~al\mbox{.}}{2020}]%
        {li2020tea}
\bibfield{author}{\bibinfo{person}{Yan Li}, \bibinfo{person}{Bin Ji},
  \bibinfo{person}{Xintian Shi}, \bibinfo{person}{Jianguo Zhang},
  \bibinfo{person}{Bin Kang}, {and} \bibinfo{person}{Limin Wang}.}
  \bibinfo{year}{2020}\natexlab{}.
\newblock \showarticletitle{TEA: Temporal Excitation and Aggregation for Action
  Recognition}. In \bibinfo{booktitle}{\emph{IEEE Conf. Comput. Vis. Pattern
  Recog.}} \bibinfo{pages}{909--918}.
\newblock


\bibitem[\protect\citeauthoryear{Lin, Li, Wang, Tai, Luo, Cui, Wang, Li, Huang,
  and Ji}{Lin et~al\mbox{.}}{2020}]%
        {lin2020fast}
\bibfield{author}{\bibinfo{person}{Chuming Lin}, \bibinfo{person}{Jian Li},
  \bibinfo{person}{Yabiao Wang}, \bibinfo{person}{Ying Tai},
  \bibinfo{person}{Donghao Luo}, \bibinfo{person}{Zhipeng Cui},
  \bibinfo{person}{Chengjie Wang}, \bibinfo{person}{Jilin Li},
  \bibinfo{person}{Feiyue Huang}, {and} \bibinfo{person}{Rongrong Ji}.}
  \bibinfo{year}{2020}\natexlab{}.
\newblock \showarticletitle{Fast Learning of Temporal Action Proposal via Dense
  Boundary Generator.}. In \bibinfo{booktitle}{\emph{AAAI}}.
  \bibinfo{pages}{11499--11506}.
\newblock


\bibitem[\protect\citeauthoryear{Lin, Gan, and Han}{Lin et~al\mbox{.}}{2019a}]%
        {tsm}
\bibfield{author}{\bibinfo{person}{Ji Lin}, \bibinfo{person}{Chuang Gan}, {and}
  \bibinfo{person}{Song Han}.} \bibinfo{year}{2019}\natexlab{a}.
\newblock \showarticletitle{TSM: Temporal Shift Module for Efficient Video
  Understanding}. In \bibinfo{booktitle}{\emph{Int. Conf. Comput. Vis.}}
\newblock


\bibitem[\protect\citeauthoryear{Lin, Liu, Li, Ding, and Wen}{Lin
  et~al\mbox{.}}{2019b}]%
        {lin2019bmn}
\bibfield{author}{\bibinfo{person}{Tianwei Lin}, \bibinfo{person}{Xiao Liu},
  \bibinfo{person}{Xin Li}, \bibinfo{person}{Errui Ding}, {and}
  \bibinfo{person}{Shilei Wen}.} \bibinfo{year}{2019}\natexlab{b}.
\newblock \showarticletitle{Bmn: Boundary-matching network for temporal action
  proposal generation}. In \bibinfo{booktitle}{\emph{ICCV}}.
  \bibinfo{pages}{3889--3898}.
\newblock


\bibitem[\protect\citeauthoryear{Lin, Zhao, Su, Wang, and Yang}{Lin
  et~al\mbox{.}}{2018}]%
        {lin2018bsn}
\bibfield{author}{\bibinfo{person}{Tianwei Lin}, \bibinfo{person}{Xu Zhao},
  \bibinfo{person}{Haisheng Su}, \bibinfo{person}{Chongjing Wang}, {and}
  \bibinfo{person}{Ming Yang}.} \bibinfo{year}{2018}\natexlab{}.
\newblock \showarticletitle{Bsn: Boundary sensitive network for temporal action
  proposal generation}. In \bibinfo{booktitle}{\emph{Proceedings of the
  European Conference on Computer Vision (ECCV)}}. \bibinfo{pages}{3--19}.
\newblock


\bibitem[\protect\citeauthoryear{Liu, Ma, Zhang, Liu, and Chang}{Liu
  et~al\mbox{.}}{2019}]%
        {liu2019multi}
\bibfield{author}{\bibinfo{person}{Yuan Liu}, \bibinfo{person}{Lin Ma},
  \bibinfo{person}{Yifeng Zhang}, \bibinfo{person}{Wei Liu}, {and}
  \bibinfo{person}{Shih-Fu Chang}.} \bibinfo{year}{2019}\natexlab{}.
\newblock \showarticletitle{Multi-granularity generator for temporal action
  proposal}. In \bibinfo{booktitle}{\emph{CVPR}}. \bibinfo{pages}{3604--3613}.
\newblock


\bibitem[\protect\citeauthoryear{Liu, Luo, Wang, Wang, Tai, Wang, Li, Huang,
  and Lu}{Liu et~al\mbox{.}}{2020}]%
        {teinet}
\bibfield{author}{\bibinfo{person}{Zhaoyang Liu}, \bibinfo{person}{Donghao
  Luo}, \bibinfo{person}{Yabiao Wang}, \bibinfo{person}{Limin Wang},
  \bibinfo{person}{Ying Tai}, \bibinfo{person}{Chengjie Wang},
  \bibinfo{person}{Jilin Li}, \bibinfo{person}{Feiyue Huang}, {and}
  \bibinfo{person}{Tong Lu}.} \bibinfo{year}{2020}\natexlab{}.
\newblock \showarticletitle{TEINet: Towards an Efficient Architecture for Video
  Recognition.}. In \bibinfo{booktitle}{\emph{AAAI}}.
  \bibinfo{pages}{11669--11676}.
\newblock


\bibitem[\protect\citeauthoryear{Mao, Ge, Shen, Tian, Wang, and Wang}{Mao
  et~al\mbox{.}}{2021}]%
        {mao2021tfpose}
\bibfield{author}{\bibinfo{person}{Weian Mao}, \bibinfo{person}{Yongtao Ge},
  \bibinfo{person}{Chunhua Shen}, \bibinfo{person}{Zhi Tian},
  \bibinfo{person}{Xinlong Wang}, {and} \bibinfo{person}{Zhibin Wang}.}
  \bibinfo{year}{2021}\natexlab{}.
\newblock \showarticletitle{TFPose: Direct Human Pose Estimation with
  Transformers}.
\newblock \bibinfo{journal}{\emph{arXiv preprint arXiv:2103.15320}}
  (\bibinfo{year}{2021}).
\newblock


\bibitem[\protect\citeauthoryear{Neimark, Bar, Zohar, and Asselmann}{Neimark
  et~al\mbox{.}}{2021}]%
        {neimark2021video}
\bibfield{author}{\bibinfo{person}{Daniel Neimark}, \bibinfo{person}{Omri Bar},
  \bibinfo{person}{Maya Zohar}, {and} \bibinfo{person}{Dotan Asselmann}.}
  \bibinfo{year}{2021}\natexlab{}.
\newblock \showarticletitle{Video transformer network}.
\newblock \bibinfo{journal}{\emph{arXiv preprint arXiv:2102.00719}}
  (\bibinfo{year}{2021}).
\newblock


\bibitem[\protect\citeauthoryear{Oneata, Verbeek, and Schmid}{Oneata
  et~al\mbox{.}}{2014}]%
        {oneata2014lear}
\bibfield{author}{\bibinfo{person}{Dan Oneata}, \bibinfo{person}{Jakob
  Verbeek}, {and} \bibinfo{person}{Cordelia Schmid}.}
  \bibinfo{year}{2014}\natexlab{}.
\newblock \showarticletitle{The lear submission at thumos 2014}.
\newblock  (\bibinfo{year}{2014}).
\newblock


\bibitem[\protect\citeauthoryear{Qing, Su, Gan, Wang, Wu, Wang, Qiao, Yan, Gao,
  and Sang}{Qing et~al\mbox{.}}{2021}]%
        {qing2021temporal}
\bibfield{author}{\bibinfo{person}{Zhiwu Qing}, \bibinfo{person}{Haisheng Su},
  \bibinfo{person}{Weihao Gan}, \bibinfo{person}{Dongliang Wang},
  \bibinfo{person}{Wei Wu}, \bibinfo{person}{Xiang Wang}, \bibinfo{person}{Yu
  Qiao}, \bibinfo{person}{Junjie Yan}, \bibinfo{person}{Changxin Gao}, {and}
  \bibinfo{person}{Nong Sang}.} \bibinfo{year}{2021}\natexlab{}.
\newblock \showarticletitle{Temporal Context Aggregation Network for Temporal
  Action Proposal Refinement}.
\newblock \bibinfo{journal}{\emph{arXiv preprint arXiv:2103.13141}}
  (\bibinfo{year}{2021}).
\newblock


\bibitem[\protect\citeauthoryear{Qiu, Yao, and Mei}{Qiu et~al\mbox{.}}{2017}]%
        {p3d}
\bibfield{author}{\bibinfo{person}{Zhaofan Qiu}, \bibinfo{person}{Ting Yao},
  {and} \bibinfo{person}{Tao Mei}.} \bibinfo{year}{2017}\natexlab{}.
\newblock \showarticletitle{Learning spatio-temporal representation with
  pseudo-3d residual networks}. In \bibinfo{booktitle}{\emph{Int. Conf. Comput.
  Vis.}} \bibinfo{pages}{5533--5541}.
\newblock


\bibitem[\protect\citeauthoryear{Shou, Chan, Zareian, Miyazawa, and Chang}{Shou
  et~al\mbox{.}}{2017}]%
        {shou2017cdc}
\bibfield{author}{\bibinfo{person}{Zheng Shou}, \bibinfo{person}{Jonathan
  Chan}, \bibinfo{person}{Alireza Zareian}, \bibinfo{person}{Kazuyuki
  Miyazawa}, {and} \bibinfo{person}{Shih-Fu Chang}.}
  \bibinfo{year}{2017}\natexlab{}.
\newblock \showarticletitle{CDC: convolutional-de-convolutional networks for
  precise temporal action localization in untrimmed videos}. In
  \bibinfo{booktitle}{\emph{CVPR}}.
\newblock


\bibitem[\protect\citeauthoryear{Shou, Wang, and Chang}{Shou
  et~al\mbox{.}}{2016}]%
        {shou2016temporal}
\bibfield{author}{\bibinfo{person}{Zheng Shou}, \bibinfo{person}{Dongang Wang},
  {and} \bibinfo{person}{Shih-Fu Chang}.} \bibinfo{year}{2016}\natexlab{}.
\newblock \showarticletitle{Temporal action localization in untrimmed videos
  via multi-stage cnns}. In \bibinfo{booktitle}{\emph{CVPR}}.
\newblock


\bibitem[\protect\citeauthoryear{Simonyan and Zisserman}{Simonyan and
  Zisserman}{2014}]%
        {two-stream}
\bibfield{author}{\bibinfo{person}{Karen Simonyan} {and}
  \bibinfo{person}{Andrew Zisserman}.} \bibinfo{year}{2014}\natexlab{}.
\newblock \showarticletitle{Two-stream convolutional networks for action
  recognition in videos}. In \bibinfo{booktitle}{\emph{NIPS}}.
  \bibinfo{pages}{568--576}.
\newblock


\bibitem[\protect\citeauthoryear{Singh and Cuzzolin}{Singh and
  Cuzzolin}{2016}]%
        {singh2016untrimmed}
\bibfield{author}{\bibinfo{person}{Gurkirt Singh} {and} \bibinfo{person}{Fabio
  Cuzzolin}.} \bibinfo{year}{2016}\natexlab{}.
\newblock \showarticletitle{Untrimmed video classification for activity
  detection: submission to activitynet challenge}.
\newblock \bibinfo{journal}{\emph{arXiv preprint arXiv:1607.01979}}
  (\bibinfo{year}{2016}).
\newblock


\bibitem[\protect\citeauthoryear{Su, Gan, Wu, Yan, and Qiao}{Su
  et~al\mbox{.}}{2020}]%
        {su2020bsn++}
\bibfield{author}{\bibinfo{person}{Haisheng Su}, \bibinfo{person}{Weihao Gan},
  \bibinfo{person}{Wei Wu}, \bibinfo{person}{Junjie Yan}, {and}
  \bibinfo{person}{Yu Qiao}.} \bibinfo{year}{2020}\natexlab{}.
\newblock \showarticletitle{BSN++: Complementary Boundary Regressor with
  Scale-Balanced Relation Modeling for Temporal Action Proposal Generation}.
\newblock \bibinfo{journal}{\emph{arXiv preprint arXiv:2009.07641}}
  (\bibinfo{year}{2020}).
\newblock


\bibitem[\protect\citeauthoryear{Tan, Tang, Wang, and Wu}{Tan
  et~al\mbox{.}}{2021}]%
        {tan2021relaxed}
\bibfield{author}{\bibinfo{person}{Jing Tan}, \bibinfo{person}{Jiaqi Tang},
  \bibinfo{person}{Limin Wang}, {and} \bibinfo{person}{Gangshan Wu}.}
  \bibinfo{year}{2021}\natexlab{}.
\newblock \showarticletitle{Relaxed Transformer Decoders for Direct Action
  Proposal Generation}.
\newblock \bibinfo{journal}{\emph{arXiv preprint arXiv:2102.01894}}
  (\bibinfo{year}{2021}).
\newblock


\bibitem[\protect\citeauthoryear{Tran, Bourdev, Fergus, Torresani, and
  Paluri}{Tran et~al\mbox{.}}{2015}]%
        {c3d}
\bibfield{author}{\bibinfo{person}{Du Tran}, \bibinfo{person}{Lubomir Bourdev},
  \bibinfo{person}{Rob Fergus}, \bibinfo{person}{Lorenzo Torresani}, {and}
  \bibinfo{person}{Manohar Paluri}.} \bibinfo{year}{2015}\natexlab{}.
\newblock \showarticletitle{Learning spatiotemporal features with 3d
  convolutional networks}. In \bibinfo{booktitle}{\emph{Int. Conf. Comput.
  Vis.}}
\newblock


\bibitem[\protect\citeauthoryear{Tran, Wang, Torresani, and Feiszli}{Tran
  et~al\mbox{.}}{2019}]%
        {CSN}
\bibfield{author}{\bibinfo{person}{Du Tran}, \bibinfo{person}{Heng Wang},
  \bibinfo{person}{Lorenzo Torresani}, {and} \bibinfo{person}{Matt Feiszli}.}
  \bibinfo{year}{2019}\natexlab{}.
\newblock \showarticletitle{Video classification with channel-separated
  convolutional networks}. In \bibinfo{booktitle}{\emph{Int. Conf. Comput.
  Vis.}} \bibinfo{pages}{5552--5561}.
\newblock


\bibitem[\protect\citeauthoryear{Tran, Wang, Torresani, Ray, LeCun, and
  Paluri}{Tran et~al\mbox{.}}{2018}]%
        {r2+1d}
\bibfield{author}{\bibinfo{person}{Du Tran}, \bibinfo{person}{Heng Wang},
  \bibinfo{person}{Lorenzo Torresani}, \bibinfo{person}{Jamie Ray},
  \bibinfo{person}{Yann LeCun}, {and} \bibinfo{person}{Manohar Paluri}.}
  \bibinfo{year}{2018}\natexlab{}.
\newblock \showarticletitle{A Closer Look at Spatiotemporal Convolutions for
  Action Recognition}. In \bibinfo{booktitle}{\emph{IEEE Conf. Comput. Vis.
  Pattern Recog.}}
\newblock


\bibitem[\protect\citeauthoryear{Vaswani, Shazeer, Parmar, Uszkoreit, Jones,
  Gomez, Kaiser, and Polosukhin}{Vaswani et~al\mbox{.}}{2017}]%
        {vaswani2017attention}
\bibfield{author}{\bibinfo{person}{Ashish Vaswani}, \bibinfo{person}{Noam
  Shazeer}, \bibinfo{person}{Niki Parmar}, \bibinfo{person}{Jakob Uszkoreit},
  \bibinfo{person}{Llion Jones}, \bibinfo{person}{Aidan~N Gomez},
  \bibinfo{person}{{\L}ukasz Kaiser}, {and} \bibinfo{person}{Illia
  Polosukhin}.} \bibinfo{year}{2017}\natexlab{}.
\newblock \showarticletitle{Attention is all you need}. In
  \bibinfo{booktitle}{\emph{Advances in neural information processing
  systems}}. \bibinfo{pages}{5998--6008}.
\newblock


\bibitem[\protect\citeauthoryear{Wang, Xiong, Lin, and Van~Gool}{Wang
  et~al\mbox{.}}{2017}]%
        {wang2017untrimmednets}
\bibfield{author}{\bibinfo{person}{Limin Wang}, \bibinfo{person}{Yuanjun
  Xiong}, \bibinfo{person}{Dahua Lin}, {and} \bibinfo{person}{Luc Van~Gool}.}
  \bibinfo{year}{2017}\natexlab{}.
\newblock \showarticletitle{Untrimmednets for weakly supervised action
  recognition and detection}. In \bibinfo{booktitle}{\emph{Proceedings of the
  IEEE conference on Computer Vision and Pattern Recognition}}.
  \bibinfo{pages}{4325--4334}.
\newblock


\bibitem[\protect\citeauthoryear{Wang, Xiong, Wang, Qiao, Lin, Tang, and
  Van~Gool}{Wang et~al\mbox{.}}{2016}]%
        {tsn}
\bibfield{author}{\bibinfo{person}{Limin Wang}, \bibinfo{person}{Yuanjun
  Xiong}, \bibinfo{person}{Zhe Wang}, \bibinfo{person}{Yu Qiao},
  \bibinfo{person}{Dahua Lin}, \bibinfo{person}{Xiaoou Tang}, {and}
  \bibinfo{person}{Luc Van~Gool}.} \bibinfo{year}{2016}\natexlab{}.
\newblock \showarticletitle{Temporal segment networks: Towards good practices
  for deep action recognition}. In \bibinfo{booktitle}{\emph{European
  conference on computer vision}}. Springer, \bibinfo{pages}{20--36}.
\newblock


\bibitem[\protect\citeauthoryear{Wu, He, Lin, Li, Gan, and Ding}{Wu
  et~al\mbox{.}}{2021}]%
        {wu2020MVFNet}
\bibfield{author}{\bibinfo{person}{Wenhao Wu}, \bibinfo{person}{Dongliang He},
  \bibinfo{person}{Tianwei Lin}, \bibinfo{person}{Fu Li},
  \bibinfo{person}{Chuang Gan}, {and} \bibinfo{person}{Errui Ding}.}
  \bibinfo{year}{2021}\natexlab{}.
\newblock \showarticletitle{MVFNet: Multi-View Fusion Network for Efficient
  Video Recognition}. In \bibinfo{booktitle}{\emph{AAAI}}.
\newblock


\bibitem[\protect\citeauthoryear{Wu, He, Tan, Chen, and Wen}{Wu
  et~al\mbox{.}}{2019a}]%
        {wu2019multi}
\bibfield{author}{\bibinfo{person}{Wenhao Wu}, \bibinfo{person}{Dongliang He},
  \bibinfo{person}{Xiao Tan}, \bibinfo{person}{Shifeng Chen}, {and}
  \bibinfo{person}{Shilei Wen}.} \bibinfo{year}{2019}\natexlab{a}.
\newblock \showarticletitle{Multi-Agent Reinforcement Learning Based Frame
  Sampling for Effective Untrimmed Video Recognition}. In
  \bibinfo{booktitle}{\emph{Int. Conf. Comput. Vis.}}
\newblock


\bibitem[\protect\citeauthoryear{Wu, He, Tan, Chen, Yang, and Wen}{Wu
  et~al\mbox{.}}{2020}]%
        {wu2020dynamic}
\bibfield{author}{\bibinfo{person}{Wenhao Wu}, \bibinfo{person}{Dongliang He},
  \bibinfo{person}{Xiao Tan}, \bibinfo{person}{Shifeng Chen},
  \bibinfo{person}{Yi Yang}, {and} \bibinfo{person}{Shilei Wen}.}
  \bibinfo{year}{2020}\natexlab{}.
\newblock \showarticletitle{Dynamic Inference: A New Approach Toward Efficient
  Video Action Recognition}. In \bibinfo{booktitle}{\emph{Proceedings of CVPR
  Workshops}}. \bibinfo{pages}{676--677}.
\newblock


\bibitem[\protect\citeauthoryear{Wu, Xiong, Ma, Socher, and Davis}{Wu
  et~al\mbox{.}}{2019b}]%
        {wu2019adaframe}
\bibfield{author}{\bibinfo{person}{Zuxuan Wu}, \bibinfo{person}{Caiming Xiong},
  \bibinfo{person}{Chih-Yao Ma}, \bibinfo{person}{Richard Socher}, {and}
  \bibinfo{person}{Larry~S Davis}.} \bibinfo{year}{2019}\natexlab{b}.
\newblock \showarticletitle{Adaframe: Adaptive frame selection for fast video
  recognition}. In \bibinfo{booktitle}{\emph{IEEE Conf. Comput. Vis. Pattern
  Recog.}} \bibinfo{pages}{1278--1287}.
\newblock


\bibitem[\protect\citeauthoryear{Xie, Sun, Huang, Tu, and Murphy}{Xie
  et~al\mbox{.}}{2018}]%
        {s3d}
\bibfield{author}{\bibinfo{person}{Saining Xie}, \bibinfo{person}{Chen Sun},
  \bibinfo{person}{Jonathan Huang}, \bibinfo{person}{Zhuowen Tu}, {and}
  \bibinfo{person}{Kevin Murphy}.} \bibinfo{year}{2018}\natexlab{}.
\newblock \showarticletitle{Rethinking spatiotemporal feature learning:
  Speed-accuracy trade-offs in video classification}. In
  \bibinfo{booktitle}{\emph{Eur. Conf. Comput. Vis.}}
\newblock


\bibitem[\protect\citeauthoryear{Xiong, Wang, Wang, Zhang, Song, Li, Lin, Qiao,
  Van~Gool, and Tang}{Xiong et~al\mbox{.}}{2016}]%
        {xiong2016cuhk}
\bibfield{author}{\bibinfo{person}{Yuanjun Xiong}, \bibinfo{person}{Limin
  Wang}, \bibinfo{person}{Zhe Wang}, \bibinfo{person}{Bowen Zhang},
  \bibinfo{person}{Hang Song}, \bibinfo{person}{Wei Li}, \bibinfo{person}{Dahua
  Lin}, \bibinfo{person}{Yu Qiao}, \bibinfo{person}{Luc Van~Gool}, {and}
  \bibinfo{person}{Xiaoou Tang}.} \bibinfo{year}{2016}\natexlab{}.
\newblock \showarticletitle{Cuhk \& ethz \& siat submission to activitynet
  challenge 2016}.
\newblock \bibinfo{journal}{\emph{arXiv preprint arXiv:1608.00797}}
  (\bibinfo{year}{2016}).
\newblock


\bibitem[\protect\citeauthoryear{Xu, Das, and Saenko}{Xu et~al\mbox{.}}{2017}]%
        {xu2017r}
\bibfield{author}{\bibinfo{person}{Huijuan Xu}, \bibinfo{person}{Abir Das},
  {and} \bibinfo{person}{Kate Saenko}.} \bibinfo{year}{2017}\natexlab{}.
\newblock \showarticletitle{R-c3d: Region convolutional 3d network for temporal
  activity detection}. In \bibinfo{booktitle}{\emph{ICCV}}.
\newblock


\bibitem[\protect\citeauthoryear{Xu, Zhao, Rojas, Thabet, and Ghanem}{Xu
  et~al\mbox{.}}{2020}]%
        {xu2020g}
\bibfield{author}{\bibinfo{person}{Mengmeng Xu}, \bibinfo{person}{Chen Zhao},
  \bibinfo{person}{David~S Rojas}, \bibinfo{person}{Ali Thabet}, {and}
  \bibinfo{person}{Bernard Ghanem}.} \bibinfo{year}{2020}\natexlab{}.
\newblock \showarticletitle{G-TAD: Sub-Graph Localization for Temporal Action
  Detection}. In \bibinfo{booktitle}{\emph{CVPR}}.
  \bibinfo{pages}{10156--10165}.
\newblock


\bibitem[\protect\citeauthoryear{Xu, Ban, Delorme, Gan, Rus, and
  Alameda-Pineda}{Xu et~al\mbox{.}}{2021}]%
        {xu2021transcenter}
\bibfield{author}{\bibinfo{person}{Yihong Xu}, \bibinfo{person}{Yutong Ban},
  \bibinfo{person}{Guillaume Delorme}, \bibinfo{person}{Chuang Gan},
  \bibinfo{person}{Daniela Rus}, {and} \bibinfo{person}{Xavier
  Alameda-Pineda}.} \bibinfo{year}{2021}\natexlab{}.
\newblock \showarticletitle{TransCenter: Transformers with Dense Queries for
  Multiple-Object Tracking}.
\newblock \bibinfo{journal}{\emph{arXiv preprint arXiv:2103.15145}}
  (\bibinfo{year}{2021}).
\newblock


\bibitem[\protect\citeauthoryear{Yang, Yang, Fu, Lu, and Guo}{Yang
  et~al\mbox{.}}{2020}]%
        {yang2020learning}
\bibfield{author}{\bibinfo{person}{Fuzhi Yang}, \bibinfo{person}{Huan Yang},
  \bibinfo{person}{Jianlong Fu}, \bibinfo{person}{Hongtao Lu}, {and}
  \bibinfo{person}{Baining Guo}.} \bibinfo{year}{2020}\natexlab{}.
\newblock \showarticletitle{Learning texture transformer network for image
  super-resolution}. In \bibinfo{booktitle}{\emph{Proceedings of the IEEE/CVF
  Conference on Computer Vision and Pattern Recognition}}.
  \bibinfo{pages}{5791--5800}.
\newblock


\bibitem[\protect\citeauthoryear{Yue-Hei~Ng, Hausknecht, Vijayanarasimhan,
  Vinyals, Monga, and Toderici}{Yue-Hei~Ng et~al\mbox{.}}{2015}]%
        {yue2015beyond}
\bibfield{author}{\bibinfo{person}{Joe Yue-Hei~Ng}, \bibinfo{person}{Matthew
  Hausknecht}, \bibinfo{person}{Sudheendra Vijayanarasimhan},
  \bibinfo{person}{Oriol Vinyals}, \bibinfo{person}{Rajat Monga}, {and}
  \bibinfo{person}{George Toderici}.} \bibinfo{year}{2015}\natexlab{}.
\newblock \showarticletitle{Beyond short snippets: Deep networks for video
  classification}. In \bibinfo{booktitle}{\emph{Proceedings of the IEEE
  conference on computer vision and pattern recognition}}.
  \bibinfo{pages}{4694--4702}.
\newblock


\bibitem[\protect\citeauthoryear{Zhao, Xiong, Wang, Wu, Tang, and Lin}{Zhao
  et~al\mbox{.}}{2017}]%
        {zhao2017temporal}
\bibfield{author}{\bibinfo{person}{Yue Zhao}, \bibinfo{person}{Yuanjun Xiong},
  \bibinfo{person}{Limin Wang}, \bibinfo{person}{Zhirong Wu},
  \bibinfo{person}{Xiaoou Tang}, {and} \bibinfo{person}{Dahua Lin}.}
  \bibinfo{year}{2017}\natexlab{}.
\newblock \showarticletitle{Temporal action detection with structured segment
  networks}. In \bibinfo{booktitle}{\emph{ICCV}}. \bibinfo{pages}{2914--2923}.
\newblock


\bibitem[\protect\citeauthoryear{Zhu, Hiller, Ehsanpour, Ma, Drummond, and
  Rezatofighi}{Zhu et~al\mbox{.}}{2021}]%
        {zhu2021looking}
\bibfield{author}{\bibinfo{person}{Tianyu Zhu}, \bibinfo{person}{Markus
  Hiller}, \bibinfo{person}{Mahsa Ehsanpour}, \bibinfo{person}{Rongkai Ma},
  \bibinfo{person}{Tom Drummond}, {and} \bibinfo{person}{Hamid Rezatofighi}.}
  \bibinfo{year}{2021}\natexlab{}.
\newblock \showarticletitle{Looking Beyond Two Frames: End-to-End Multi-Object
  Tracking Using Spatial and Temporal Transformers}.
\newblock \bibinfo{journal}{\emph{arXiv preprint arXiv:2103.14829}}
  (\bibinfo{year}{2021}).
\newblock


\end{thebibliography}

\appendix

\end{document}


\title{Supplementary Material for  \\ Temporal Action Proposal Generation with Transformers}

\begin{teaserfigure}
	\begin{center}
	\includegraphics[width=0.9\textwidth]{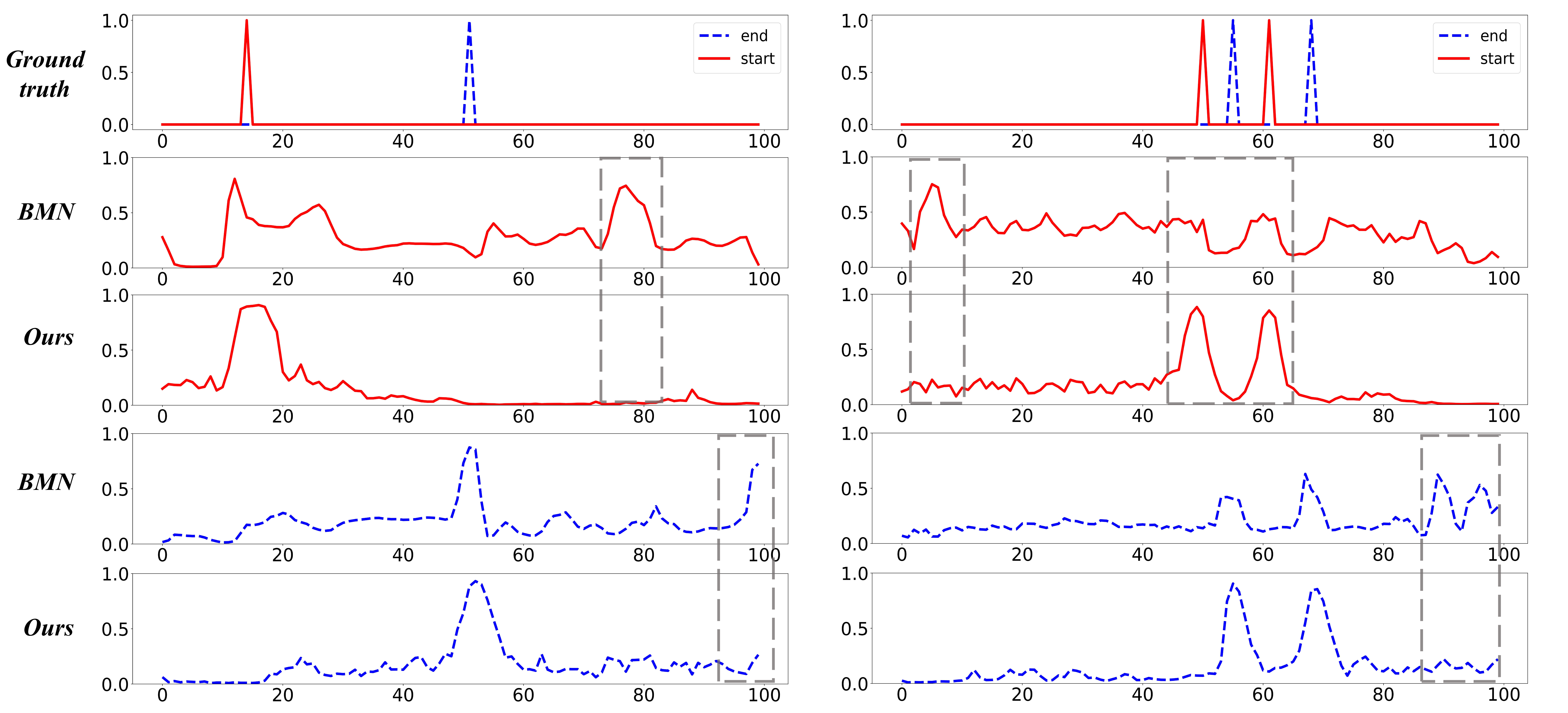}
	\end{center}
	\caption{Visualization examples of starting and ending probability sequences on ActivityNet1.3.}
   	\label{fig:visual_sup_1}
\end{teaserfigure}

\maketitle

\section{Label assignment}
The method of label assignment in each loss functions are described in detail below.
For Boundary Transformer, we need to generate temporal boundary label sequence $G_s,G_e \in R^T$ 
Following \cite{lin2019bmn}, for a ground-truth action instance ${\varphi}_g = (t_s,t_e)$ with duration $d_g = t_e - t_s$ in annotation set $\Psi_w$, we denote its starting and ending regions as $r_S = [t_s - {d_g}/{10},t_s + {d_g}/{10}]$ and$r_E = [t_e - {d_g}/{10},t_e + {d_g}/{10}]$ separately. 
Then, for a temporal location $t_n$ within $F_w$, we denote its local region as $r_{t_n} = [t_n - d_f/2,t_n + d_f/2]$, where $d_f = t_n - t_{n-1}$ is the temporal interval between two locations.
Then we calculate overlap ratio $IoR$ of $r_{t_n}$ with $r_S$ and $r_E$ separately, and denote maximum $IoR$ as $g^s_{t_n}$ and $g^e_{t_n}$ separately, where $IoR$ is defined as the overlap ratio with ground truth proportional to the duration of this region.
Thus we can generate $G_{S,w} ={\{ g^s_{t_n}\}}^{l_w}_{n=1}$ and  $G_{E,w} ={\{ g^e_{t_n}\}}^{l_w}_{n=1}$ as label of Boundary Transformer.
For Proposal Transformer, we need to generate \textbf{Sparse Proposal sequence} label $G_C \in R^L$, where $L$is number of sparse proposals.
For a proposal $\varphi_{i} \in P$, we calculate its Intersection-over-Union $(IoU)$ with all $\varphi_g$ in $\Psi_w$, and denote the maximum $IoU$ as $g^c_{i}$.
Thus we can generate $G_c = {\{ g^c_{i}\}}^{L}_{i=1}$ as label of Proposal Transformer.

\section{More visualization examples}
Figure~\ref{fig:visual_sup_1} illustrates two visualization examples and comparison with classical method BMN~\cite{lin2019bmn}.
The example on the left shows that the BTR module can significantly reduce the adverse effects of boundary noise. The other example indicates that BTR can enhance the boundary awareness. These two remarkable advantages benefit from the ability of BTR to capture the long-term frame-level dependencies.

\begin{figure*}[!t]
	\begin{center}
	\includegraphics[width=1\textwidth]{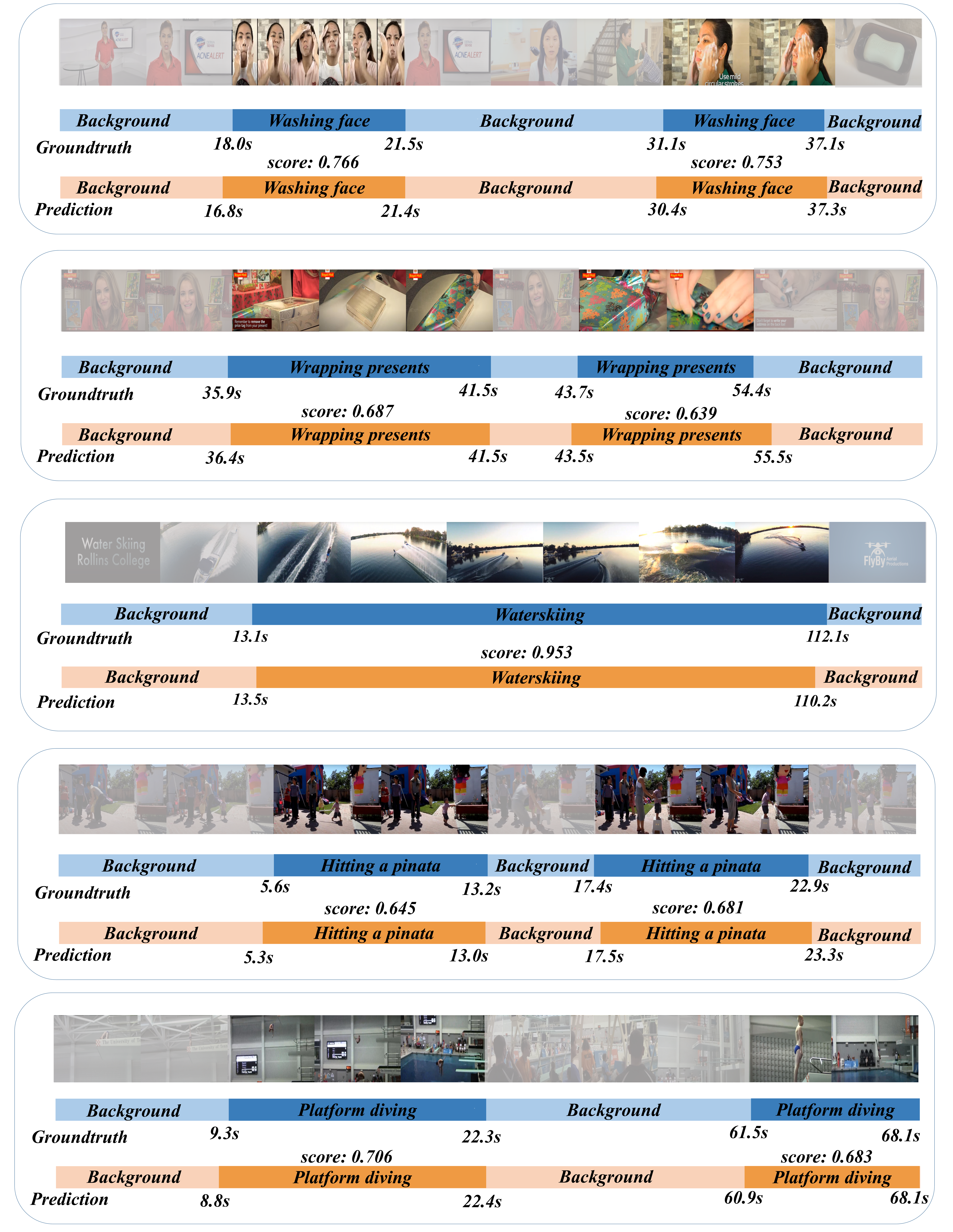}
	\end{center}
	\caption{Visualization examples of generated proposals on ActivityNet1.3.}
   	\label{fig:visual_sup_2}
\end{figure*}

Figure~\ref{fig:visual_sup_2} illustrates more visualization examples on ActivityNet1.3.
The generated proposals of our method with the highest $k$ scores are visualized in each video, where $k$ is the number of ground truth.
The scenarios of all examples are relatively complicated.
In the first and second videos which contain complex action scenes, our method predicts the various scenes correctly as a complete action. For instance, the action of wrapping presents contains various scenes including putting the present in the wrapping paper, packing the present and so on. 
The correct predictions show that our method understands the inter-frame relation and inter-proposal relation properly.
In the third and fourth videos, where the background frames and action frames are of the similar scene, our top-2 proposals are successfully aligned with the positions of ground truth action instances. 
Our method is able to identify the noisy snippets even when they have very similar appearance to the action.
In the bottom video, which contains complex background scenes, our model correctly understands the semantic information and identify the true action. For instance, the background of the platform diving action contains various scenes including audience cheering, the diving platform scene without competition going on and so on. 

\bibliographystyle{ACM-Reference-Format}
\bibliography{sample-base}
